%% file: sn-article_round1.tex
\documentclass[default,iicol]{sn-jnl}% Default with double column layout
\usepackage{amsmath}
\usepackage{diagbox}
\usepackage{graphics}
\usepackage{epsfig}
\usepackage{threeparttable}
\usepackage{xspace}
\usepackage{siunitx}
\usepackage{subfigure}
\usepackage{graphicx}
\usepackage{amssymb}
\usepackage{threeparttable}
\usepackage{color}
\usepackage[normalem]{ulem}
\usepackage{multirow}
\usepackage{float}
\usepackage{amsfonts}

\usepackage{bm}
\usepackage{array}
\PassOptionsToPackage{table}{xcolor}
\definecolor{mygray}{RGB}{200,200,200}
\definecolor{reda}{RGB}{255,0,0}
\definecolor{redb}{RGB}{217,148,143}
\definecolor{myyellow}{RGB}{190,144,0}
\definecolor{mygreen}{RGB}{0,136,51}
\definecolor{myblue}{RGB}{0,102,204}
\usepackage{colortbl}

\usepackage{pifont}

\makeatletter
\newcommand{\thickhline}{%
    \noalign {\ifnum 0=`}\fi \hrule height 1pt
    \futurelet \reserved@a \@xhline
}
\makeatother

\jyear{2021}%

\theoremstyle{thmstyleone}%
\theoremstyle{thmstyletwo}%

\theoremstyle{thmstylethree}%

\begin{document}
\title[Article Title]{Adaptive Multi-source Predictor for Zero-shot Video Object Segmentation}
%%=============================================================%%
%% Prefix	-> \pfx{Dr}
%% GivenName	-> \fnm{Joergen W.}
%% Particle	-> \spfx{van der} -> surname prefix
%% FamilyName	-> \sur{Ploeg}
%% Suffix	-> \sfx{IV}
%% NatureName	-> \tanm{Poet Laureate} -> Title after name
%% Degrees	-> \dgr{MSc, PhD}
%% \author*[1,2]{\pfx{Dr} \fnm{Joergen W.} \spfx{van der} \sur{Ploeg} \sfx{IV} \tanm{Poet Laureate} 
%%                 \dgr{MSc, PhD}}\email{iauthor@gmail.com}
%%=============================================================%%
\author[1]{\fnm{Xiaoqi} \sur{Zhao}}\email{zxq@mail.dlut.edu.cn}\equalcont{These authors contributed equally to this work.}
\author[1]{\fnm{Shijie} \sur{Chang}}\email{csj@mail.dlut.edu.cn}\equalcont{These authors contributed equally to this work.}
\author[1]{\fnm{Youwei} \sur{Pang}}\email{lartpang@mail.dlut.edu.cn}
\author[1]{\fnm{Jiaxing} \sur{Yang}}\email{jx.yang@mail.dlut.edu.com}
\author*[1]{\fnm{Lihe} \sur{Zhang} }\email{zhanglihe@dlut.edu.cn}
\author[1]{\fnm{Huchuan} \sur{Lu}}\email{lhchuan@dlut.edu.cn}

\affil[1]{\orgdiv{Dalian University of Technology, Dalian, China}}

%\affil[2]{\orgdiv{Peng Cheng Laboratory, China}}

%%==================================%%
%% sample for unstructured abstract %%
%%==================================%%

\abstract{
Static and moving objects often occur in real-life videos. 
Most video object segmentation methods only focus on extracting and exploiting motion cues to perceive moving objects. 
Once faced with the frames of static objects, the moving object predictors may predict failed results caused by uncertain motion information, such as low-quality optical flow maps. 
Besides, different sources such as RGB, depth, optical flow and static saliency can provide useful information about the objects. However, existing approaches only consider either the RGB or RGB and optical flow. 
In this paper, we propose a novel adaptive multi-source predictor for zero-shot video object segmentation (ZVOS).
In the static object predictor, the RGB source is converted to depth and static saliency sources, simultaneously.
In the moving object predictor, we propose the multi-source fusion structure. 
First, the spatial importance of each source is highlighted with the help of the interoceptive spatial attention module (ISAM). 
Second, the motion-enhanced module (MEM) is designed to generate pure foreground motion attention for improving the representation of static and moving features in the decoder. 
Furthermore, we design a feature purification module (FPM) to filter the inter-source incompatible features. By using the ISAM, MEM and FPM, the multi-source features are effectively fused.
In addition, we put forward an adaptive predictor fusion network (APF) to evaluate the quality of the optical flow map and fuse the predictions from the static object predictor and the moving object predictor in order to prevent over-reliance on the failed results caused by low-quality optical flow maps. 
Experiments show that the proposed model outperforms the state-of-the-art methods on three challenging ZVOS benchmarks. 
And, the static object predictor precisely predicts a high-quality depth map and static saliency map at the same time.}


\maketitle
\section{Introduction}\label{sec:intro}
Zero-shot Video Object Segmentation (ZVOS) aims to automatically separate primary foreground object/objects from their background in a video without human annotation for any testing frames. It has attracted a lot of interests due to the wide application scenarios such as autonomous driving, video surveillance and video editing. With the development of deep convolutional neural networks (CNN), the CNN-Based ZVOS methods dominate this field. 
According to the way of capturing the foreground or moving objects, ZVOS methods can be divided into interframe-based~\cite{WCS,AGNN,COSNet,AGS,EPO,PDB} and optical flow-based methods~\cite{GateNet,MP,SFL,MATNet,FSNet,AMCNet,RTNet}. In this paper, we build an optical flow-based ZVOS method by utilizing the optical flow to depict motion information of video objects.
\begin{figure}[t]
\includegraphics[width=\linewidth]{
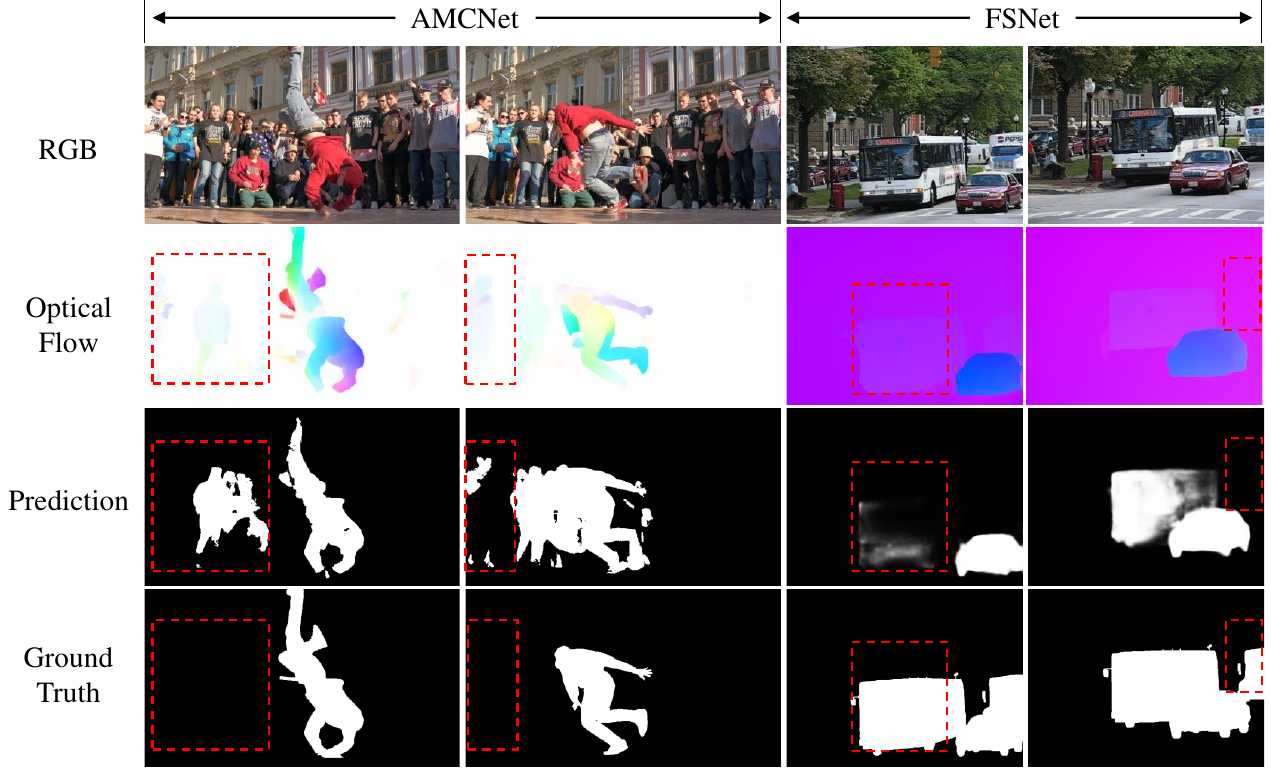}
\centering
\caption{Failure cases in AMCNet~\cite{AMCNet} and FSNet~\cite{FSNet}. These two respective sequences ($breakdance$ and $cars$) are selected from the  DAVIS$_{16}$~\cite{davis16} and FBMS~\cite{FBMS} dataset, respectively. 
These difficult samples have obvious interference from low-quality optical flow maps.}
\label{fig:Figure1}
\end{figure}

\begin{figure}[t]
\includegraphics[width=\linewidth]{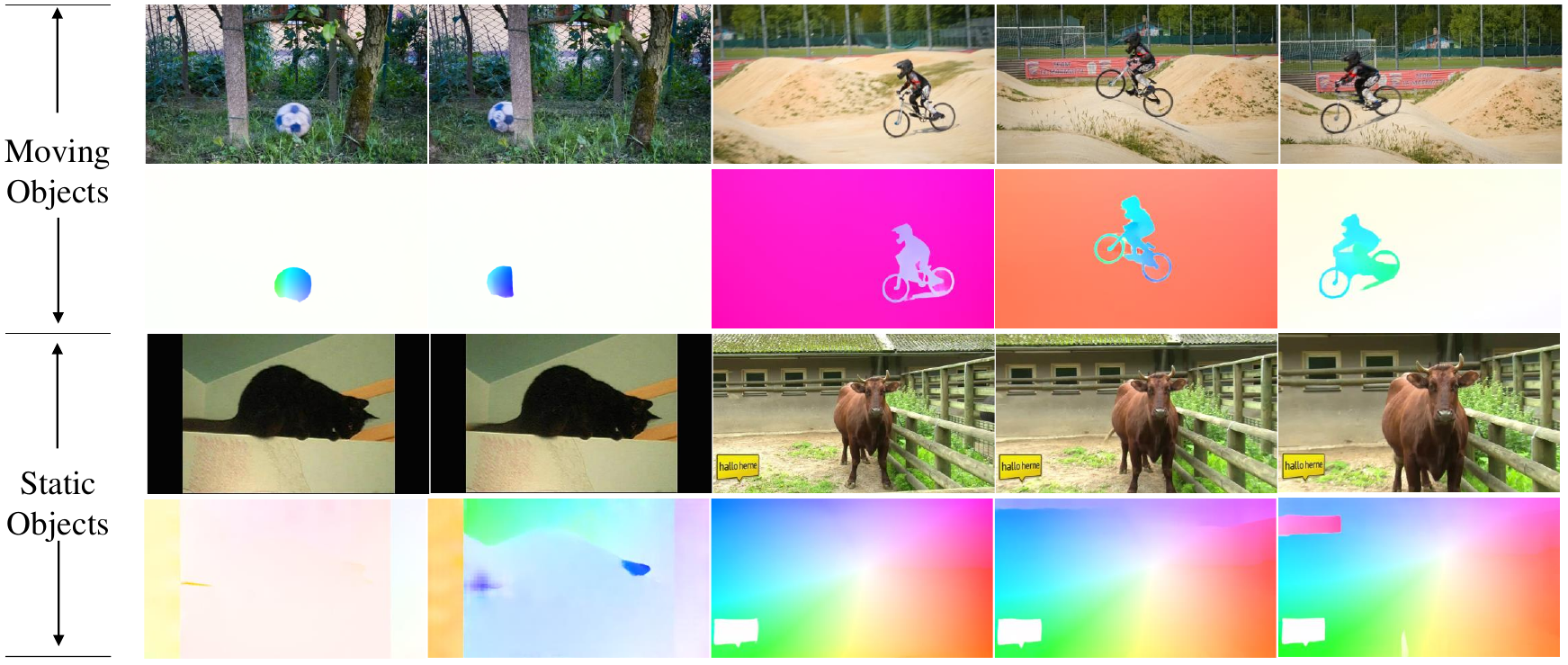}\\
        \centering
        \caption{Some pairs of RGB and optical flow maps in static and moving frames. The static  frames $cat$ and $cow$ are randomly selected from the Youtube-Objects~\cite{youtube-objects}. The moving  frames $bmx-bumps$ and $soccerball$ are randomly selected from the  DAVIS$_{16}$~\cite{davis16}.  }
\label{fig:Figure2}
\end{figure}

\begin{figure*}
    \includegraphics[width=\linewidth]{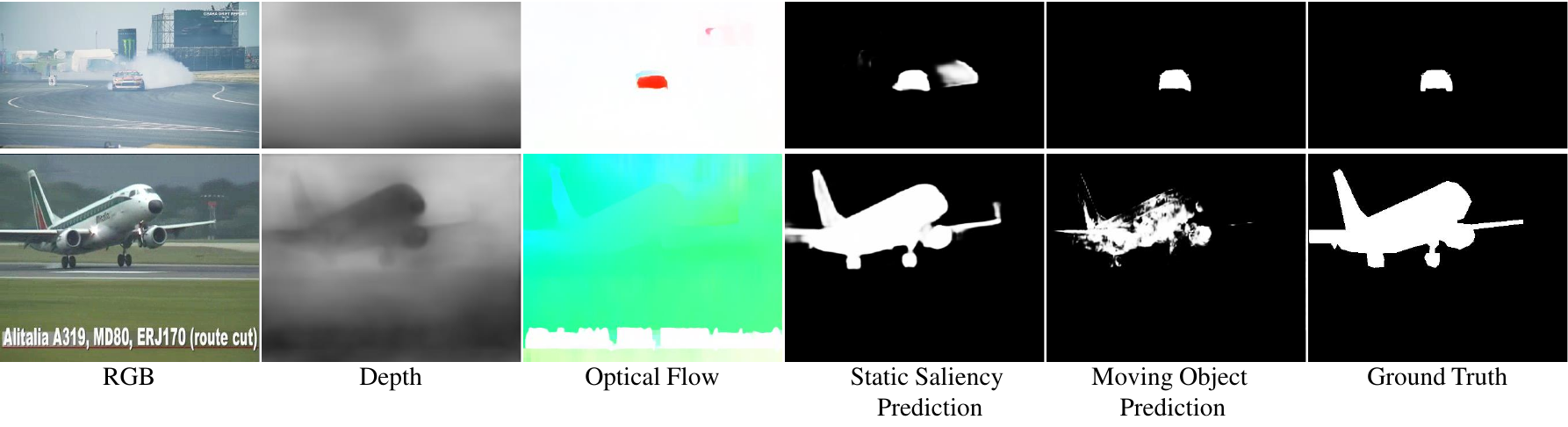}
    \centering
    \caption{Visual results of different sources. Samples $drift-chicane$ and $aeroplane$ are randomly selected from the DAVIS$_{16}$~\cite{davis16} and Youtube-Objects~\cite{youtube-objects}, respectively.     } 
    \label{fig:Figure3}
\end{figure*}

High-quality optical flow maps indeed provide significant position information of moving objects.
Many methods~\cite{MATNet,FSNet,AMCNet,RTNet} focus on deeply integrating the optical flow features at multiple levels. However, over-reliance on optical flow information can cause negative effects. 
As shown in Fig.~\ref{fig:Figure1}, both AMCNet~\cite{AMCNet} and FSNet~\cite{FSNet} produce failed predictions guided by low-quality optical flow maps. 
As we know, a video sequence is made up of a series of static images.
%The process of observing objects is from static to dynamic. 
If the object in a video no longer moves or moves very slowly, the task of static object segmentation is equivalent to salient object detection. When a object is moving in the scene, object segmentation can be promoted by optical flow because it contains the patterns of objects, surfaces and edges. As can be seen from Fig.~\ref{fig:Figure2}, the  frames with static objects have low quality optical flow maps compared to those of moving objects. It is necessary to add a static object predictor into the optical flow-based ZVOS branch, thereby improving the robustness of video object predictors. 
In addition, the depth map can also provide useful complementary information for segmentation tasks, such as RGB-D semantic segmentation~\cite{PA-RGBD-SS,DA-RGBD-SS,SS-RGBD-SS} and  saliency segmentation~\cite{SSLSOD,DANet_RGBDSOD,HDFNet_RGBDSOD}. 
{As shown in Fig.~\ref{fig:Figure3}, we visualize the various sources mentioned previously. It can be seen that the depth map and static saliency provide the $aeroplane$ with a better appearance contour information while the optical flow map accurately shows the position information of $drift-chicane$ without background interference. Actually, the RGB, optical flow, depth and static saliency show good complementarity in the representation of position and appearance for video object segmentation.}
However, all previous ZVOS methods only focus on the RGB or RGB and optical flow, other sources are neglected. 
In Fig.~\ref{fig:Figure3}, we can see that neither the static prediction nor the moving object prediction consistently show good results across all frames. The quality of optical flow map and depth map strongly affects the performance of the moving object predictor and the static object predictor, respectively. How to adaptively choose the appropriate predictor and even merge them to yield better results is also a key challenge.

Motivated by these observations, we propose a novel adaptive multi-source predictor for ZVOS. 
\textbf{\textit{Firstly}}, we design a simple multi-task static predictor, which aims to predict the depth map and static saliency map from a single RGB image, simultaneously. It adopts the FPN~\cite{FPN} structure with one encoder and two decoders. The encoder extracts the RGB features, while the decoders infer the depth and static saliency features, respectively.
\textbf{\textit{Secondly}}, we construct the moving object predictor based on multi-source fusion strategy.
Specifically, the interoceptive spatial attention module (ISAM) effectively combines the feature maps provided by four kinds of sources (i.e, RGB, depth, optical flow and static saliency). The ISAM can adaptively perceive the importance of each source features in their spatial positions compared to other sources, thereby preserving the source-specific information in the fused features. Our ISAM structure is simple yet flexible, which can be also easily equipped to the multi-task static predictor for achieving cross-modal feature fusion. 
To take full advantage of position and edge patterns in optical flow, we put forward the motion-enhanced module (MEM) to further improve the representation capability of both static and moving features. 
Since multiple sources contain some mutual interference effects, we build the feature purification module (FPM) to filter out the incompatible information. With the help of ISAM, MEM and FPM, the moving object can be segmented precisely. 
\textbf{\textit{Lastly}}, we design an adaptive predictor fusion network (APF) to evaluate the objectness of the optical flow and fuse the results from static object segmentation and moving object segmentation through the generated matching weight, thereby avoiding the prediction failure caused by relying on either predictor.

Our main contributions can be summarized as:
\begin{itemize}
\item[$\bullet$] 
We present a novel solution and new insight for zero-shot video object segmentation by extending the traditional single video predictor mode to a complete dual predictors framework towards both static and moving objects.
\item[$\bullet$] 
We utilize multi-source information to extract rich appearance and motion features, including RGB, depth, optical flow and static saliency.
\item[$\bullet$] 
We design a set of multi-source fusion components, which contains the interoceptive spatial attention module, the motion-enhanced module and the feature purification module. This design can effectively integrate complementary features, thus helping the model to focus on the salient or moving object regions.
\item[$\bullet$] 
We propose an adaptive predictor fusion (APF) network to evaluate two pairs of matching degrees between input sources and predictions, thereby fusing the predictions 
%according to an adaptive weight 
and outputting a better segmentation result than either one from the single predictor.   
\item[$\bullet$] 
Experimental results indicate that the proposed method significantly surpasses the existing state-of-the-art algorithms on three popular ZVOS benchmarks DAVIS$_{16}$, Youtube-Objects and FBMS. Besides, the multi-task static predictor has comparable performance on RGB-D salient object detection (RGB-D SOD), which can predict high-quality depth maps and salient object segmentation at the same time.
\item[$\bullet$] 
We conduct thorough ablation studies for APF on both ZVOS and RGB-D SOD tasks to show the general capability of evaluating the quality of the optical flow and depth map. 
\end{itemize}

\textit{Compared with the MM version~\cite{MSAPS} of this work, the following extensions are made. 
\textbf{\uppercase\expandafter{\romannumeral1})} In the multi-task static predictor, we modify the previous independent decoders into the interaction decoders by embedding the ISAM structure, which can improve both static salient object segmentation and depth estimation.
\textbf{\uppercase\expandafter{\romannumeral2})}
We build a motion enhancement module (MEM) to fully mine position and shape patterns in optical flow, which can enhance the perception of moving objects in the multi-source fusion stage.
\textbf{\uppercase\expandafter{\romannumeral3})} To improve the fault tolerance rate of the previous automatic predictor selection (APS) network, we replace the hard selection strategy with the current adaptive predictor fusion (APF) strategy. Moreover, the annotations used in APF are jointly generated by five segmentation metrics instead of the single MAE score measured in APS. In this way, the final prediction from APF has the advantages of low bias and high comprehensiveness. 
\textbf{\uppercase\expandafter{\romannumeral4})} We further provide more implementation details and thorough ablation studies at qualitative and quantitative aspects.
\textbf{\uppercase\expandafter{\romannumeral5})} We report much more extensive experimental results in RGB-D salient object detection and depth estimation that demonstrate the superiority of the multi-task static predictor.
\textbf{\uppercase\expandafter{\romannumeral6})} We perform in-depth analyses for the APF and further carry forward the spirit of evaluating the quality of the depth map in RGB-D SOD methods.    }

\section{Related Work}\label{sec:related_work}

\subsection{Zero-shot Video Object Segmentation}\label{subsec:zvos}

Different from one-shot video object segmentation (OVOS) which is given one or more annotated frames (the first frame in general), zero-shot video object segmentation (ZVOS) aims to automatically detect the target object without any human definition. Many CNN-based methods~\cite{PDB, MotAdapt, EPO, AGS, COSNet, AGNN} are proposed to utilize the inter-frame relationship to capture rich context and enable more complete understanding of video content. For instance, the recurrent neural network is used to capture temporal information in~\cite{PDB,AGS}. Lu~\textit{et al.}~\cite{COSNet} take a pair of frames as input and learn their correlations by using the co-attention mechanism. Wang~\textit{et al.}~\cite{AGNN} propose an attended graph neural network and perform recursive message passing to mine the underlying high-order correlations.

In addition, optical flow itself can provide important motion information. Benefiting from some outstanding optical flow estimation methods~\cite{PWC,MaskFlowNet,RAFT}, optical flow map can be easily obtained and applied to ZVOS. Tokmakov \textit{et al.}~\cite{MP} only use the optical flow map as the input and build a fully convolutional network to segment the moving object. But this map can not provide sufficient appearance information compared to the RGB input. In~\cite{LVO,SFL,UVOS-Bilateral,MATNet,FSNet,AMCNet}, two parallel streams are built to extract features from the RGB image and optical flow map, which are further fused in the decoder to predict the segmentation results. The MATNet~\cite{MATNet} achieves the transition of attended motion features to enhance appearance learning at each convolution stage. The FSNet~\cite{FSNet} presents a full-duplex strategy that ensures mutual restraint between appearance and motion information. The AMCNet~\cite{AMCNet} build a multi-modality co-attention network for fusing appearance and motion information. The RTNet~\cite{RTNet} exploits the intraframe contrast and motion cues to segment primary objects from the videos. However, the aforementioned optical flow-based methods depend on the optical flow map heavily when achieving appearance and motion feature fusion in multiple layers of the network. Once the optical flow map quality is very low, it is bound to result in very serious interference. To address this issue, we put forward an adaptive predictor fusion network to judge the effectiveness of the optical flow-based predictor.

\subsection{Static Object Segmentation}\label{subsec:sos}
Static object segmentation (SOS) aims to segment nearly stationary foreground from images or videos. 
When facing the static object without the guidance of motion information, SOS can also be viewed as the salient object detection (SOD) task. 
General CNN-based RGB SOD methods focus on the attention mechanism~\cite{BMPM,PAGRN,PFA,GateNet}, edge enhancement~\cite{R3Net,BASNet,EGNet,F3Net} and multi-scale localization~\cite{U2Net,MINet,PoolNet}. 
With the application of depth sensors, RGB-D segmentation gains lots of research interest. 
Most RGB-D segmentation methods~\cite{SSLSOD,SPNet_RGBDSOD,RD3D_RGBDSOD,DCF_RGBDSOD,JLDCF_RGBDSOD,DSA2F_RGBDSOD} adopt the two-stream structure to extract rich features from both RGB and depth, and then design diverse cross-modal fusion modules to provide the decoder with RGB-D features integrating complementary information.
Some approaches~\cite{DASNet_RGBDSOD,CoNet_RGBDSOD} use depth maps only in the training stage and achieve a depth-free inferring. Compared to depth-based methods, depth-free methods not only save parameters from an extra depth encoder but also avoid absorbing failure depth maps directly from the depth sensor. Therefore, we follow the depth-free style and design a simple yet effective multi-task network as the static predictor. 
\begin{figure*}
    \includegraphics[width=\linewidth]{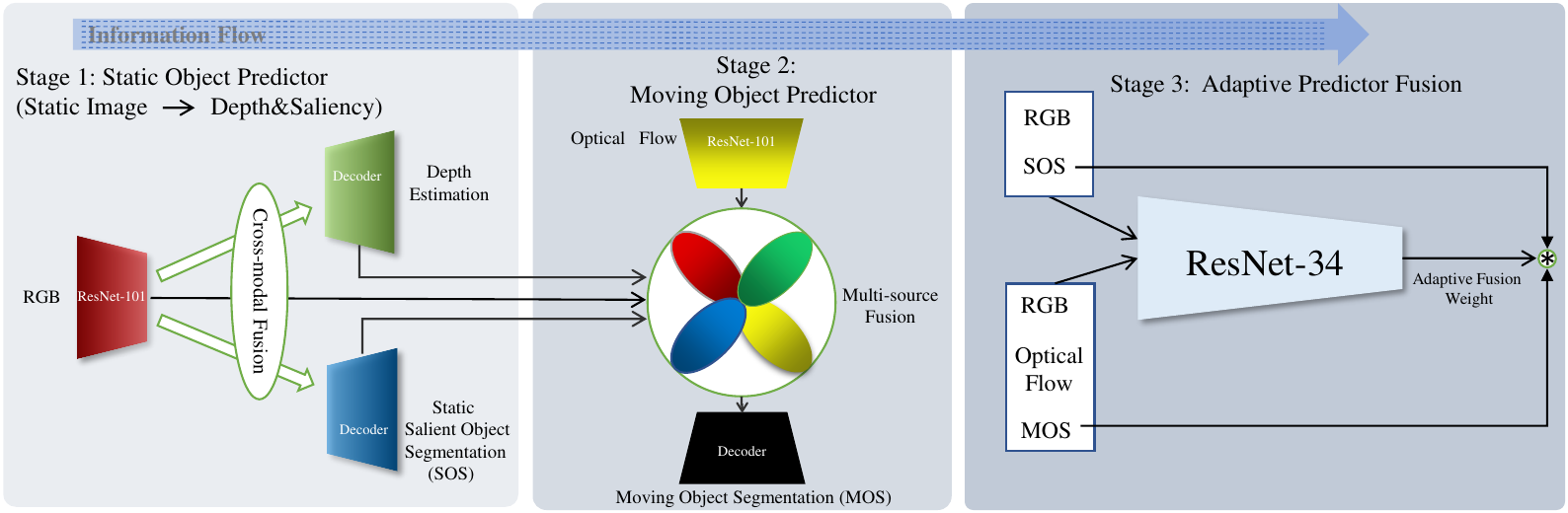}
    \centering
    \caption{Network pipeline of the ZVOS task. It consists of three stages: static object predictor, moving object predictor and adaptive predictor fusion.
  {The first stage network has two-fold function: 1) It used to generate features of RGB, depth and static saliency for the second stage. 2) It can produce static salient object segmentation (SOS). 
    The second stage network achieves multi-source fusion and yields moving object segmentation prediction (MOS).} 
    The third stage network fuses the predictions from the static object predictor and moving object predictor as the final output.} 		
    \label{fig:Figure4}
\end{figure*}
\subsection{Multi-source Information}\label{subsec:multi-source}

RGB refers to three channels of red, green and blue. This standard includes almost all colors that human vision can perceive, and is one of the most widely used color systems. Many computer vision tasks, such as classification, object detection, object tracking and semantic segmentation use the RGB image as the main input source. However, only relying on the RGB source is difficult to handle some complex environments such as low-contrast objects, which share similar appearances to the background. 
Benefiting from Microsoft Kinect and Intel RealSense devices, depth information can be conveniently obtained. Moreover, the stable geometric structures depicted in the depth map are robust against the changes of illumination and texture, which can provide important supplementary information for handling complex scenarios. 
Therefore, many RGB-D methods are applied in different tasks, such as RGB-D semantic segmentation~\cite{Gate-RGBD-SS,PA-RGBD-SS,DA-RGBD-SS,SS-RGBD-SS,LS-RGBD-SS}, RGB-D salient object detection, and RGB-D tracking \cite{rgbd-tracking1,rgbd-tracking2, rgbd-tracking3}. 

In addition, as an important computer vision task, salient object detection also needs to delineate the location and contour information of salient objects in a scene. The objects to be segmented by ZVOS usually have visual saliency, thus salient object detection can provide crucial cues for ZVOS. Many methods~\cite{GateNet,BASNet,DSS} are proposed to solve this basic vision task. For the video related task, motion information is a key attribute. How to effectively capture motion prior has received much attention, where optical flow estimation is an active research branch. Recently, many CNN-based methods~\cite{SFN,PWC,liteflownet,VCN,RAFT} utilize iterative refinement strategy to improve the performance of optical flow estimation. However, to the best of our knowledge, all ZVOS approaches only utilize one or two sources (RGB or RGB and optical flow) and neglect other sources which have lots of complementary information. In this work, we aim to take full advantage of the useful information from the aforementioned multiple sources and weed out conflicting semantic information among them to solve the zero-shot video object segmentation.

\section{Methodology}\label{sec:methodology}
Overall, we segment the objects of motion and appearance saliency within video frames by using a three-stage pipeline, as shown in Fig.~\ref{fig:Figure4}. In the first stage, a network capable of achieving simultaneously depth estimation and static salient object segmentation (SOS) is built, which only feeds on static images. In the second stage, a network able to fuse features from different sources (RGB image, depth, static saliency, and optical flow) is engineered to achieve moving object segmentation (MOS), with an interoceptive spatial attention module (ISAM), a motion-enhanced module (MEM) and a feature purification module (FPM) at its core. In the third stage, an adaptive predictor fusion (APF) network is applied to fuse SOS and MOS outputs according to their respective confidence scores. We will detail the three stages and show how they are coordinated.

\subsection{Static Object Predictor}\label{subsec:Static_predictor}

\begin{figure}[t]
\includegraphics[width=\linewidth]{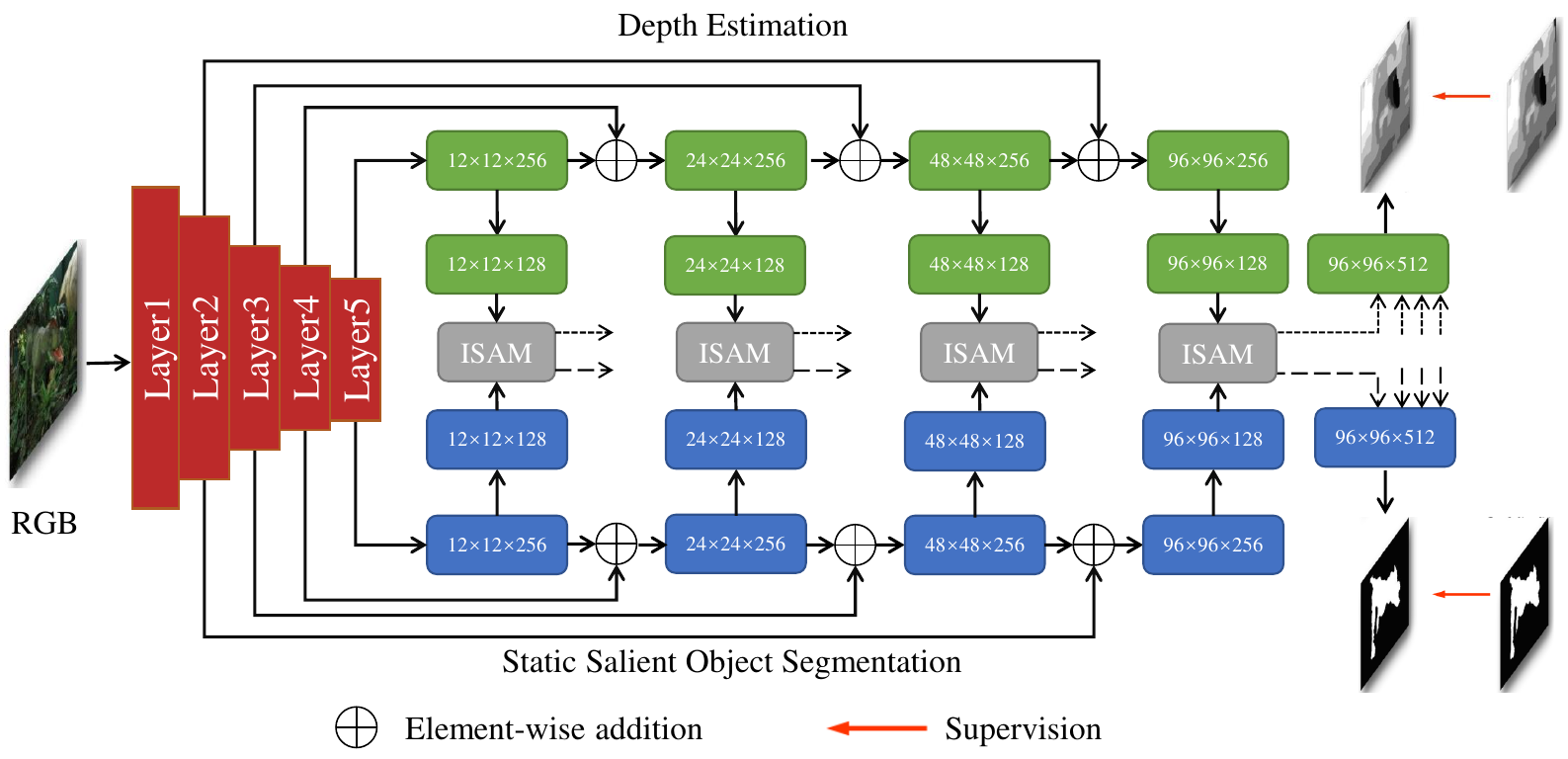}
        \centering
        \caption{Illustration of the static predictor network.}
\label{fig:Figure5}
% \vspace{-5mm}
\end{figure}

The static object predictor in this stage (adjusted from FPN~\cite{FPN}) has a uniencoder-bidecoder structure, as illustrated in Fig.~\ref{fig:Figure5}, in which the encoder is fed with static images and the decoders will predict depth and static saliency maps, respectively. 
The encoder uses tail-cast ResNet-101~\cite{ResNet} to accommodate the requirement of our task (fully-convolutional). 
The two-stream decoders have five upsampling stages
to gradually restore resolution of the embedding, during which the features from each encoder layer are fused into their corresponding decoder stages via the self-explaining skip-connection.
Depth map contains natural contrast information, which provides useful guidance for static saliency segmentation. Saliency map provides a fine foreground segmentation and its internal consistency is helpful to avoid the checkerboard pattern in depth estimation. To integrate both complementary information, we apply the interoceptive spatial attention module (ISAM) to finish cross-modal feature fusion and embed them in each decoder block. The details of ISAM refer to the~\ref{subsec:Moving Object Predictor}. 
In addition, we apply foreground object ground truth and depth annotation to supervise their outputs, respectively. For the depth stream, we follow the related works~\cite{depth3,depth4,depth5} to adopt the combination loss of L1 and SSIM~\cite{SSIM}. For another, we use weighted BCE loss~\cite{BCE} and weighted IoU loss as in~\cite{F3Net}. Such double supervision strategy will drive the transformation of source semantics, namely, RGB source to depth source and static saliency source. Note that compared to the methods directly adopting existing depth estimation and saliency detection networks, our multi-task design will save training and inference time and a number of parameters.

To be specific, our static predictor network will produce: (1) the RGB feature map $F_{rgb}^i (i\in {\{1, 2, 3, 4, 5\}})$ is extracted from the $i$th level of the encoder; (2) the depth feature map $F_{d}^i(i\in {\{2, 3, 4, 5\}})$ from the decoder of depth estimation; (3) the static saliency feature map $F_{ss}^i(i\in {\{2, 3, 4, 5\}})$ from the decoder of salient object segmentation; (4) the depth map and the static saliency map $M_{sos}$ above three kinds of feature maps will be fed into the second stage network together for moving object segmentation.

\subsection{Moving Object Predictor}\label{subsec:Moving Object Predictor}
\begin{figure*}
\includegraphics[width=\linewidth]{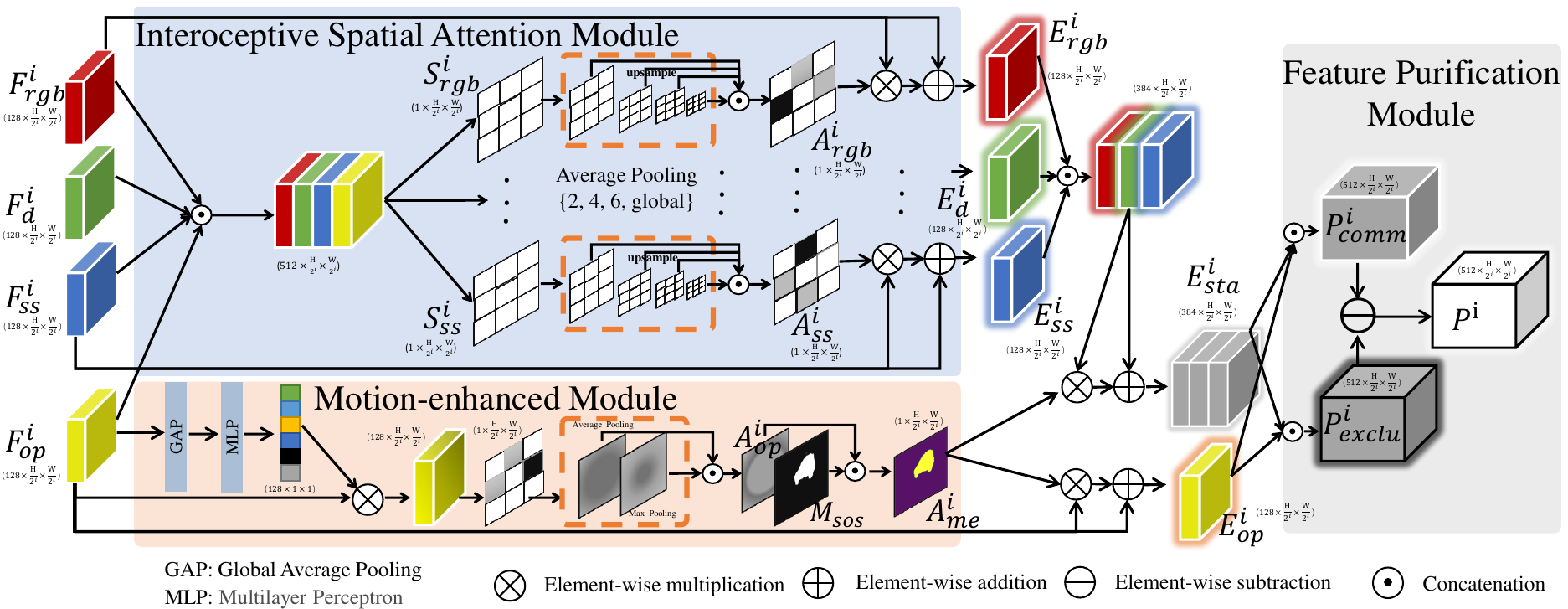}
        \centering
        \caption{Detailed diagram of interoceptive spatial attention module (ISAM), motion-enhanced module (MEM) and feature purification module (FPM). {We represent  some important feature maps with the shapes of C $\times$  H/2$^{i}$ $\times$  W/2$^{i}$, (C is the channel number, $i \in \left \{1, 2, 3, 4, 5 \right \}$ indexes different levels.) to make the whole process understandable.}}
\label{fig:Figure6}
\end{figure*}

The moving object predictor consists of interoceptive spatial attention modules (ISAM), motion-enhanced modules (MEM), feature purification modules (FPM), an encoder and a decoder. Note that the encoder in this stage is only used to extract motion feature maps $F_{op}^i (i\in {\{1, 2, 3, 4, 5\}})$ from optical flow, which is calculated by feeding temporal close frames into the off-the-shelf RAFT Net~\cite{RAFT}. Thus, the features of the four sources are well-prepared. The interoceptive spatial attention module enhances the spatial saliency of the feature maps of the static sources (RGB, depth, static saliency) by computing spatial-attention maps. The motion-enhanced module further emphasizes the guidance of motion semantic localization by channel-attention mechanism. The feature purification module calculates the difference between features containing inter-source common information and mutually-exclusive information to filter out incompatible contexts. Combination of ISAM, MEM and FPM completes effective multi-source fusion. Their structures are shown in Fig.~\ref{fig:Figure6}.

\textbf{Interoceptive Spatial Attention Module (ISAM)}: Within the network, ISAM first concatenates all the source feature maps to generate a single-channel interoceptive feature map for each static source, as follows:
\begin{equation} 
S_{src}^i={Conv}_1({Conv}_{512}(Cat(F_{rgb}^i, F_{d}^i, F_{ss}^i, F_{op}^i))),
\end{equation}
where $src \in {\{rgb, d, ss\}}$ describes the identities of RGB image
source, depth source and static saliency source,
respectively. The ${Conv}_{512}(\cdot{})$ and ${Conv}_{1}(\cdot{})$ operations refer to the $1\times1$ convolutions with 512 output channels and $3\times3$ convolutions with one output channel, respectively, by which the independent source features are correlated. ${Cat}(\cdot{})$ is the concatenation operation along channel axis. Then, we compute an interoceptive attention map with multi-scale information. This process is formulated as follows:
\begin{equation} 
{A}_{src}^i=Sig({Conv}_1(Cat(Up(MP(S_{src}^i))))),
\end{equation}
where $MP$ denotes a group of pooling operations along the channel dimension with scale $\{1, 2, 4, 6, global\}$, ${Up}(\cdot{})$ is the bilinear interpolation to upsample the features map to the same size as $S_{src}^i$ and ${Sig}(\cdot{})$ is  the element-wise sigmoid function. The ${A}_{src}^i$ is used to enhance each source feature as follows:
\begin{equation} 
{E}_{src}^i={F}_{src}^i+Conv({F}_{src}^i\otimes{}{A}_{src}^i),
\end{equation}
where $Conv(\cdot{})$ denotes $3\times3$ convolutions, $\otimes{}$ represents element-wise multiplication, and ${E}_{src}^i$ is the enhanced feature at the $i$th level.

\textbf{Motion-Enhanced Module (MEM)}: The optical flow features is the key cues to provide the moving objects with motion information. To fully exploit the gain it brings to the multi-source fusion, we enhance the semantic localization while maintaining its spatial information. 
{In Fig.~\ref{fig:Figure6}, we thoroughly illustrate the structure of motion-enhanced module. First, global average pooling operation and two fully connected layers are used to enhance the channel attention for the optical flow feature $F_{op}^i$. Next, we use $7\times7$ convolution with one output channel and a sigmoid activation function to generate a spatial attention map. Then, average pooling and max pooling can improve the sensitivity of the spatial attention map to texture and background information, respectively. Once the comprehensive channel and spatial attention map ${A}_{op}^i$ is well prepared, we combine the static saliency map $M_{sos}$ with ${A}_{op}^i$ to generate the motion-enhanced map ${A}_{me}^i$. Finally, ${A}_{me}^i$ will separately act on the aggregated static features and optical flow features to generate the enhanced features ${E}_{sta}^i$ and ${E}_{op}^i$.}

\textbf{Feature Purification Module (FPM)}: Following ISAM and MEM, FPM at every level takes the concatenation of ${E}_{sta}^i$ and ${E}_{op}^i$ to obtain the fused feature. There is the incompatibility problem when fusing multi-source features. Direct combination ${P}_{comm}^i$ is not sufficient to dilute incompatible components. Therefore, we construct an auxiliary feature, namely ${P}_{exclu}^i$. The fusion process is formulated as:
\begin{equation} 
{P}^i={P}_{comm}^i-{P}_{exclu}^i,
\end{equation}
where ${P}_{comm}^i$ and ${P}_{exclu}^i$ can be represented as:
\begin{equation}
\begin{aligned}
{P}_{comm}^i={Conv}_{512}(Cat({E}_{sta}^{i}, {E}_{op}^{i})),
\\
{P}_{exclu}^i={Conv}_{512}(Cat({E}_{sta}^{i}, {E}_{op}^{i})).
\end{aligned}
\end{equation}
${P}_{comm}^i$ and ${P}_{exclu}^i$ have different convolution parameters although their formulas are the same. The rationale behind FPM is that the subtraction  and the MOS-used supervision will force features ${P}_{comm}^i$ and ${P}_{exclu}^i$ to represent the common and mutually-exclusive contexts of the four sources, respectively. After preparing ${P}^i$ at each level, we gradually combine all of them to generate the final prediction $M_{mos}$ of moving object segmentation by the plain decoder.

\begin{figure}[!t]
\includegraphics[width=\linewidth]{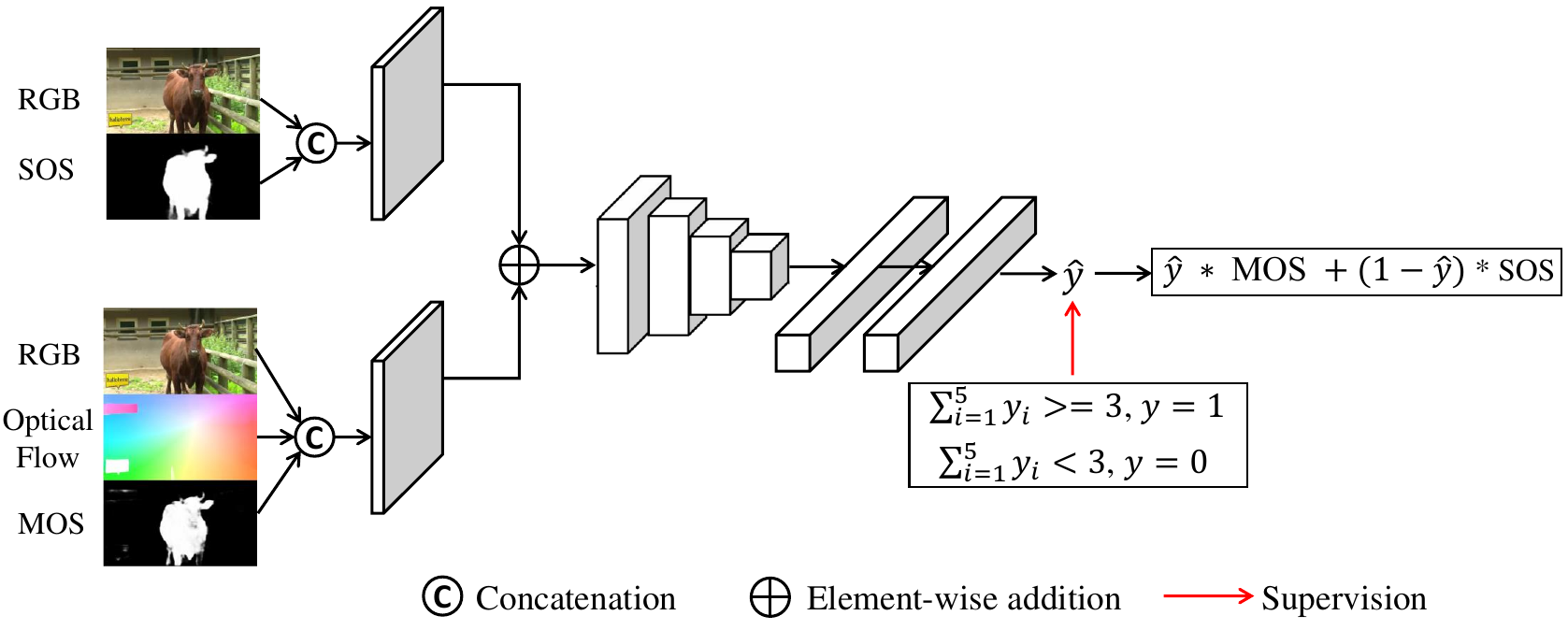}\\
        \centering
        \caption{The overall architecture of the adaptive predictor fusion network.}
\label{fig:Figure7}
\end{figure}

\subsection{Adaptive Predictor Fusion}\label{subsec:Automatic Predictor Selection}

In the third stage, the proposed Adaptive Predictor Fusion Network (APF) will produce the confidence score measured from SOS and MOS, and then play the role of an adaptive weight to fuse and output the final segmentation result. 
As shown in Fig.~\ref{fig:Figure7}, the overall network can also be viewed as solving a binary classification problem. 
We adopt a lightweight ResNet-34~\cite{ResNet} in order to reduce the amount of parameters and also for easier training. During the forward propagation, two inputs are required: (1) the concatenation of RGB and the SOS result; (2) the concatenation of RGB, optical flow map and the MOS result. For mathematical tractability, we use $rs$ and $rm$ to represent them, respectively. We replace the first layer of the encoder with the convolution of $64$ output channels, so that it can perceive the two inputs of different sizes. We initialize the parameters of the added convolutional layers with He’s method~\cite{PRelu}. After the first layer, we integrate the two-stream feature outputs by element-wise addition. The fused feature maps are fed into the other layers of ResNet-34 pretrained on ImageNet~\cite{ImageNet}, to alleviate the over-fitting problem to some extent. The process of the score prediction can be formulated as follows:
\begin{equation}
    \hat{y}=APF(rs, rm;\theta{}),
\end{equation}
where $\theta{}$ is the learnable parameters and $\hat{y}$ is the adaptive score. 
To enable APF to have comprehensive discrimination capabilities at the structure and detail levels, we utilize five popular segmentation metrics, mean similarity~\cite{davis16} ($\mathcal{J}$ $\uparrow$), mean boundary accuracy~\cite{davis16} ($\mathcal{F}$ $\uparrow$), S-measure~\cite{S-m} ($S_{m}$ $\uparrow$), maximum E-measure~\cite{E-m} ($E_{m}$ $\uparrow$) and mean absolute error~\cite{MAE} ($\mathcal{M}$ $\downarrow$), to dynamically generate corresponding sub-annotation $y_{i}$ ($i \in \left \{1, 2, 3, 4, 5 \right \}$: 
\begin{equation}\label{equ:11}
\centering
\left\{\begin{matrix}
\Psi(M_{sos}, G) < \Psi(M_{mos},G),  y_{i\neq5} = 1,  y_{i=5} = 0 \\
\Psi(M_{sos}, G) >= \Psi(M_{mos},G),  y_{i\neq5} = 0,  y_{i=5} = 1\\
    \end{matrix}\right.
\end{equation} 
where $\Psi$ denotes the above metric operation. 
If the performance of $M_{mos}$ dominates more metrics, we set the annotation $y$ to $1$, otherwise $y$ is set to $0$:
\begin{equation}\label{equ:16}
\centering
    \left\{\begin{matrix}
    \sum_{i=1}^5y_{i} >=3,  y = 1\\
  \sum_{i=1}^5y_{i} <3,  y = 0.  \\
    \end{matrix}\right.
\end{equation}
Finally, we adopt the binary cross entropy loss, which can be computed as:
 \begin{equation}\label{equ:17}
 \begin{split}
    \mathcal{L}_{bce} = -(ylog\hat {y}+(1-y)log(1-\hat{y})).
 \end{split}
\end{equation}
Supervised by the dynamic binary classification annotation, the APF network can learn the predictor selection rule that a lower score represents a higher confidence of the SOS and a lower degree of matching among MOS, optical flow and RGB. On the contrary, a higher score means a higher confidence of the MOS and a lower degree of matching between SOS and RGB. Therefore, the APF network implements the function of evaluating the effectiveness of object-information contained in the optical flow for ZVOS.
Once the binary classification network has learned this scoring pattern, the final output of APF can be formulated as:
\begin{equation}\label{equ:18}
    M_{apf}=\hat{y}*M_{mos}+(1-\hat{y})*M_{sos}.
\end{equation}

Compared with the strategy of rigidly selecting SOS or MOS by relying on the predicted score with an implicit 0.5 binarization in the MM version~\cite{MSAPS}, we implement a soft fusion strategy to generate the final prediction $M_{apf}$, which provides some redundancy for the accuracy of the classification network to avoid steep performance drops due to classification failures.

\section{Experiments}\label{sec:experiments}
In this section, we first compare our method with the state-of-the-art methods on three popular zero-shot video object segmentation (ZVOS) datasets. Then we conduct ablation studies to verify the effectiveness of components including multi-source features, interoceptive spatial attention module (ISAM), motion-enhanced module (MEM), feature purification module (FPM) and adaptive predictor fusion (APF) in our model. Finally, we extend the static predictor and adaptive predictor fusion network on several popular RGB-D salient object detection (SOD) datasets to show the generalization.

\subsection{Datasets and Metrics}\label{sec:metrics}

For the zero-shot video object segmentation task, we evaluate our method on three popular datasets: {DAVIS}$_{16}$~\cite{davis16}, YouTube-Objects~\cite{youtube-objects} and FBMS~\cite{FBMS}. \textbf{\textit{{DAVIS}$_{16}$}}~\cite{davis16} is one of the most popular benchmark datasets for video object segmentation task. It consists of 50 high-quality video sequences (30 for training and 20 for validation) in total. And most videos do not have special scenes such as dramatic changes in the background or almost no movement in the foreground. Usually, the high-quality optical flow map can be obtained via optical flow networks. We use two standard metrics: mean similarity $\mathcal{J}$ and mean boundary accuracy $\mathcal{F}$~\cite{davis16} to measure the segmentation results (The higher are better).
\textbf{\textit{YouTube-Objects}}~\cite{youtube-objects} is a large dataset of 126 web videos with 10 semantic object categories and more than $20,000 frames$. This dataset contains many unconventional videos, such as dramatically changing backgrounds, slowly moving or stationary objects, objects only moving in the depth dimension. The quality of the optical flow map is usually low in these video sequences. Following most methods~\cite{MATNet,WCS,AGNN,COSNet,AGS,PDB,AMCNet,RTNet}, we only use the mean region similarity $\mathcal{J}$ metric to measure the performance.
\textbf{\textit{FBMS}}~\cite{FBMS} is composed of 59 video sequences with ground truth annotations provided in a subset of the frames. We use the training set to train the automatic predictor selection network and evaluate our method on the validation set, which consists of 30 sequences. Since it is designed for their specific purposes, FBMS is not often used for the ZVOS task. We just use it for a fair comparison.

For the RGB-D salient object detection task, we evaluate the static predictor on five popular datasets: NLPR~\cite{NLPR}, NJUD~\cite{NJU2K}, SIP~\cite{SIP}, STERE~\cite{STERE} and DUTLF-D~\cite{DUTLF-D}. For fair evaluation, we follow two settings as most previous methods~\cite{SPNet_RGBDSOD,HDFNet_RGBDSOD,BBSNet_RGBDSOD,JLDCF_RGBDSOD,CoNet_RGBDSOD,DASNet_RGBDSOD}. Some~\cite{SPNet_RGBDSOD,BBSNet_RGBDSOD,JL-DCF_RGBDSOD,SPNet,DASNet_RGBDSOD} use $1,485$ samples from the NJUD and $700$ samples from the NLPR as the training set and the remaining samples in these datasets are used for testing. The others~\cite{DSA2F_RGBDSOD,DCF_RGBDSOD,HDFNet_RGBDSOD,DANet_RGBDSOD,CoNet_RGBDSOD} adopt the combination of DUTLF-D~\cite{DUTLF-D} + NJUD~\cite{NJU2K} + NLPR~\cite{NLPR} to form $2,185$ pairs of RGB-D training sets. We adopt several widely used metrics for quantitative evaluation: F-measure~\cite{colorcontrast_Fm} ($F_{\beta}$), mean absolute error~\cite{MAE} (MAE, $\mathcal{M}$), S-measure~\cite{S-m} ($S_{\alpha}$), E-measure~\cite{E-m} ($E_{\xi}$). The lower value is better for the $\mathcal{M}$ and the higher is better for others. Besides, our static predictor can predict high-quality depth maps, we adopt the following metrics used by previous works for evaluating depth estimation: Root Mean Square Error (RMSE, RMSE (log)), absolute relative error (AbsRel), squared relative error (SqRel) and depth accuracy at various thresholds $1.25$, $1.25^{2}$ and $1.25^{3}$.

\begin{table}[!t]
  \centering
  \caption{Summary of the video and image data used in different methods for training each segmentation network.}
  \begin{threeparttable}
   \resizebox{0.9\linewidth}{!}{
    \setlength\tabcolsep{5pt}
    \renewcommand\arraystretch{1}
   \input{table/scales_training}
}
\end{threeparttable}
\label{tab:scales_training}
\end{table}

\subsection{Implementation Details}\label{subsec4-2}

Our model is implemented based on Pytorch~\cite{pytorch} and trained on a PC with an RTX 3090 GPU. The input sources are all resized to $384\times384$ and all the three stages use the mini-batch of size 4 for training. First, we use some RGB-D saliency datasets~\cite{NLPR,NJUD,SIP,RGBD135,STERE,DUTLF-D} to train the static predictor network. 
Once the first stage of training is finished, we adopt the DAVIS$_{16}$ training set to train the multi-source fusion network. In the process, the parameters of the static predictor are frozen, and we just train the moving object predictor.
We use a total of $8,398$ labeled samples including more than $2,000$ frames video data and $6,300$ RGB-D SOD data in the first and second phases. The depth map is obtained by the depth camera.
In addition, we expect that there are low-quality and high-quality optical flow videos for training in the third stage, while most of the optical flow maps in the {DAVIS}$_{16}$ dataset are high-quality, it is not suitable for this training. Therefore, we use more than $2,300$ frames from the {DAVIS}$_{16}$ and FBMS training set to train the APF. 
{In Tab.~\ref{tab:scales_training}, we summarize the scales of training data used in different segmentation networks. In a word, the scale of the used annotations in our training phase is comparable to that of these competitors.}

We adopt some data augmentation techniques in each stage to avoid over-fitting: horizontally random flip, random rotate, random brightness, saturation and contrast. 
In the first and second stages, we use the AdamW optimizer~\cite{AdamW} with a betas of ($0.9$, $0.999$) and a weight decay of $0.01$. The learning rate is set to $0.00006$ and later use the ``poly'' policy~\cite{poly} with the power of $0.9$ as a means of adjustment. 
In the third stage, we use the SGD optimizer with a momentum of $0.9$, a weight decay of $0.0005$ and a learning rate of $0.001$ with ``poly'' adjustment policy. The source code can be available at \textcolor{red}{\url{https://github.com/Xiaoqi-Zhao-DLUT/Multi-Source-APS-ZVOS}}.

\subsection{Performance on ZVOS}\label{subsec:performance_vos}
\noindent\textbf{Quantitative Results.}
 We compare the proposed method with the state-of-the-art ZVOS methods. Tab.~\ref{tab:Table1} shows  performance comparison results in terms of the mean~$\mathcal{J}$ and mean~$\mathcal{F}$ on \textbf{\textit{DAVIS$_{16}$}}. It can be seen that ours can consistently outperform other approaches under the two metrics. Compared to the second best~\& the optical flow-based method (RTNet), our method achieves an important improvement of 1.5\% and 3.2\% in terms of mean $\mathcal{J}$ and mean $\mathcal{F}$, respectively. 
 Moreover, the model size of Ours vs. RTNet is $0.54$G vs. $1.03$G. Therefore, our method has significant advantages in terms of accuracy and model efficiency. Tab.~\ref{tab:Table2} shows the results on \textbf{\textit{Youtube-Objects}}~\cite{youtube-objects}. It can be seen that our method achieves the best performance in six of the ten categories. Notably, our method has a significant performance improvement of $5.5$\% in terms of mean $\mathcal{J}$ compared to the optical flow-based method AMCNet ($75.0$ vs. $71.1$). As mentioned by Zhou \textit{et al.}~\cite{MATNet}, Youtube-Objects has many video sequences containing slowly moving and/or visually indistinct objects. Both will result in inaccurate estimation of optical flow. Therefore, the existing optical flow-based methods can not perform very well on this dataset. With the help of our adaptive predictor fusion network, this problem can be solved well. In order to make a comprehensive comparison with the previous methods, we also show the results on \textbf{\textit{FBMS}}~\cite{FBMS}, as shown in Tab.~\ref{tab:Table3}. Our method still achieves the best performance compared to the others.

\begin{figure*}
    \includegraphics[width=\linewidth]{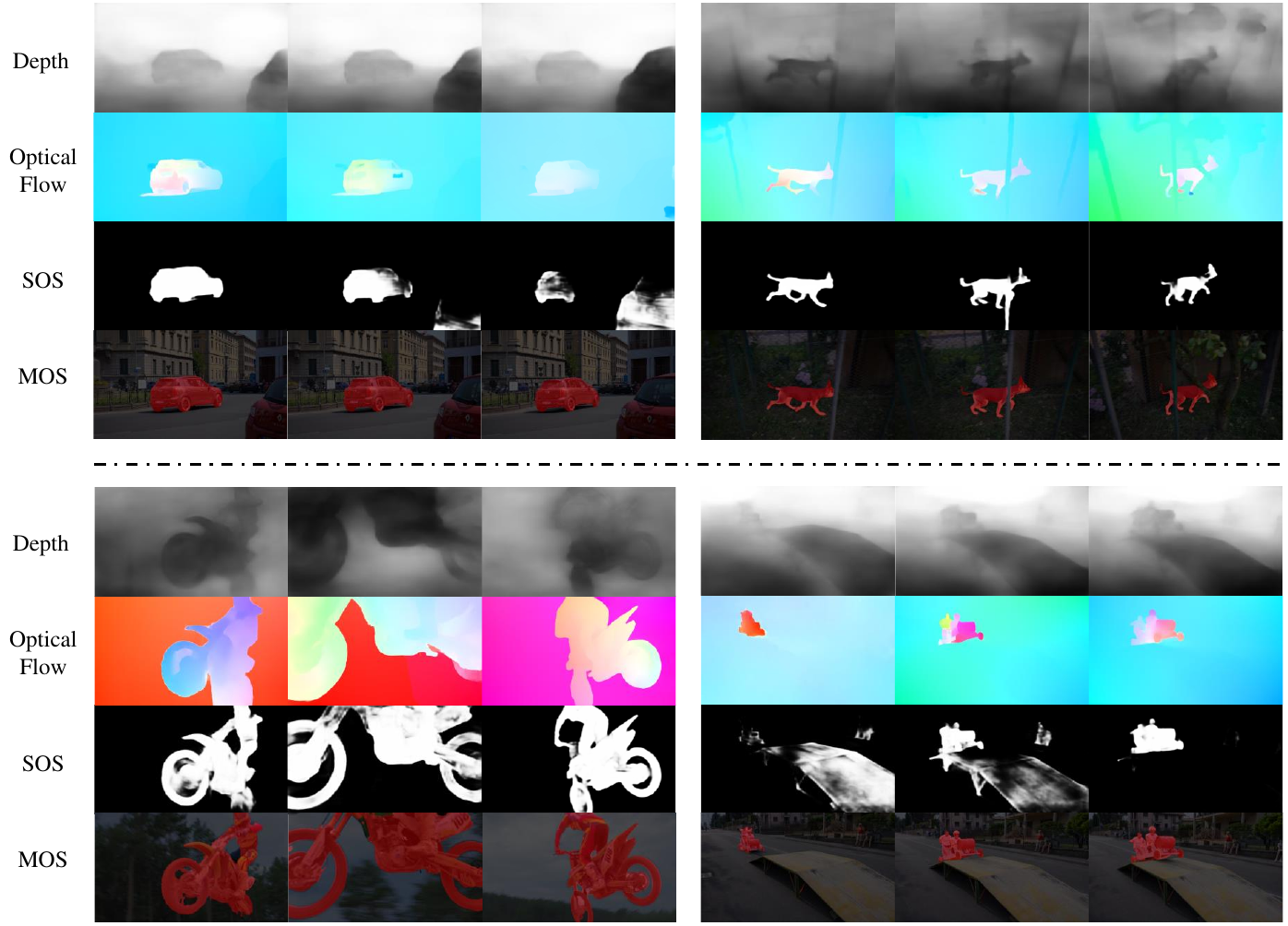}\\ %,height=0.3\linewidth
    \centering
    \caption{Qualitative results on four sequences $car-roundabout$, $libby$, $motocross-jump$ and $soapbox$ of DAVIS$_{16}$.} 		
    \label{fig:Figure8}
\end{figure*}

\noindent\textbf{Qualitative Results.}
Fig.~\ref{fig:Figure8} illustrates visual results of the proposed algorithm on the challenge video sequences $car-roundabout$, $libby$, $motocross-jump$ and $soapbox$ of DAVIS$_{16}$. We can see that each source provides rich location and appearance information. Moreover, depth map can supplement extra contrast information, and the high-quality optical flow can provide clear motion information. Notably, none of these sources can dominate the final prediction, and all of the source characteristics must be integrated to achieve high-precision video object segmentation.

\begin{table*}[htbp]
  \centering
  \caption{Quantitative comparison on the DAVIS$_{16}$~\cite{davis16} validation set. The best result for each metric is highlighted in \textbf{bold}.}
  \begin{threeparttable}
   \resizebox{\linewidth}{!}{
    \setlength\tabcolsep{5pt}
    \renewcommand\arraystretch{1}
   \input{table/davis16_comparison}
}
\end{threeparttable}
\label{tab:Table1}
\end{table*}

\begin{table*}[htbp]
  \centering
  \caption{Quantitative results of each category on the Youtube-Objects~\cite{youtube-objects} in terms of mean $\mathcal{J}$. We show the average performance for each of the $10$ categories, and the final row gives an average over all the videos.}
  \begin{threeparttable}
   \resizebox{\linewidth}{!}{
    \setlength\tabcolsep{5pt}
    \renewcommand\arraystretch{1}
   \input{table/youtube_comparison}
}
\end{threeparttable}
\label{tab:Table2}
\end{table*}

\begin{table}[htbp]
  \centering
  \caption{Quantitative results on the FBMS~\cite{FBMS} dataset.}
  \begin{threeparttable}
   \resizebox{\linewidth}{!}{
    \setlength\tabcolsep{2pt}
    \renewcommand\arraystretch{1}
   \input{table/fbms_comparison}
}
\end{threeparttable}
\label{tab:Table3}
\end{table}

\begin{table}[htbp]
  \centering
  \caption{Ablation study on the validation set of DAVIS$_{16}$.}
  \begin{threeparttable}
   \resizebox{\linewidth}{!}{
    \setlength\tabcolsep{2pt}
    \renewcommand\arraystretch{1}
   \input{table/ablation_study_davis16}
}
\end{threeparttable}
\label{tab:Table4}
\end{table}

\begin{table}[htbp]
  \centering
  \caption{{Quantitative comparison among ISAM, MEM and Transformer in terms of efficiency (training memory, module parameters) and accuracy (mean $\mathcal{J}$, mean $\mathcal{F}$)   on the validation set of DAVIS$_{16}$.}}
  \begin{threeparttable}
   \resizebox{\linewidth}{!}{
    \setlength\tabcolsep{2pt}
    \renewcommand\arraystretch{1}
   \input{table/ablation_study_davis16_replace_transformer}
}
\end{threeparttable}
\label{tab:Table4_1}
\end{table}

\subsection{Ablation Study}\label{subsec:ablation_study}
We detail the contribution of each component to the overall network. Because the optical flow maps usually are high quality on the DAVIS$_{16}$. We perform an ablation study on it to investigate the effect of multi-source fusion. Specifically, we first verify the effectiveness of each source for moving object segmentation (MOS). Next, we show the benefits of the interoceptive spatial attention module (ISAM), motion-enhanced module (MEM) and feature purification module (FPM) in the multi-source fusion network, respectively. Finally, we evaluate the performance of the adaptive predictor fusion (APF) on both DAVIS$_{16}$ and Youtube-Objects. 

\subsubsection{Effectiveness of Multi-source Features}\label{subsubsec:multi-source-features}
We quantitatively show the benefit of each source in multi-source feature inputs of Tab.~\ref{tab:Table4}.
We take the FPN with the only RGB feature inputs as the baseline. First, the depth features, static saliency features and optical flow features are added to the baseline network, respectively. It can be seen that the combination of RGB source and other sources has a significant improvement compared to the baseline, with the gain of $5.9\%$ and $5.3\%$ in terms of mean $\mathcal{J}$ and mean $\mathcal{F}$. 
In the process, multi-source features complement each other.

\begin{figure}[!t]
\includegraphics[width=\linewidth]{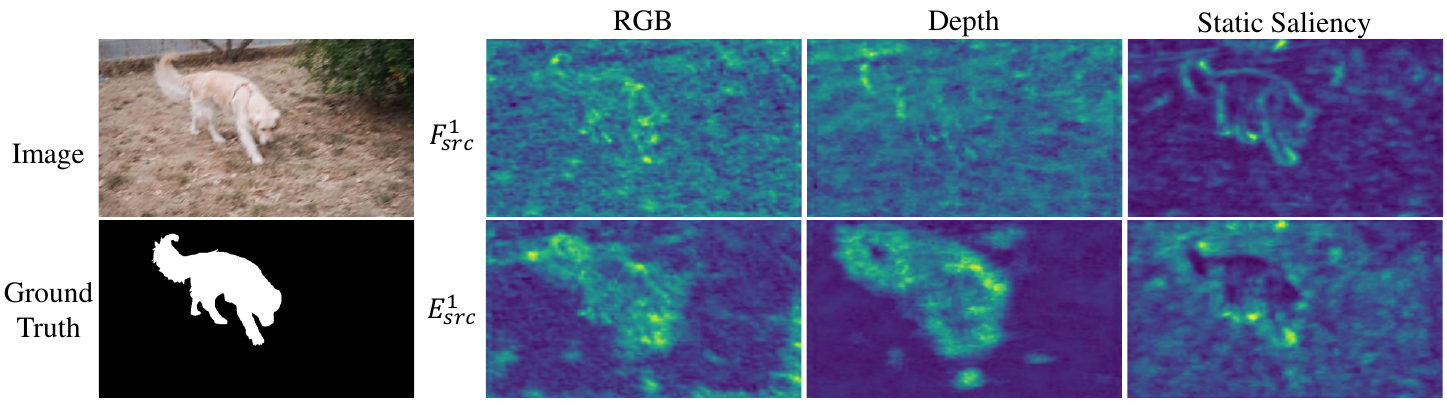}\\
        \centering
        \caption{Visual results  of  ${F}_{src}^1$ and ${E}_{src}^1$ for showing the effect of the interoceptive spatial attention module. Sample $dog$ is randomly selected from the DAVIS$_{16}$~\cite{davis16}.}
\label{fig:Figure9}
% \vspace{-5mm}
\end{figure}

\begin{figure}[!t]
\includegraphics[width=\linewidth]{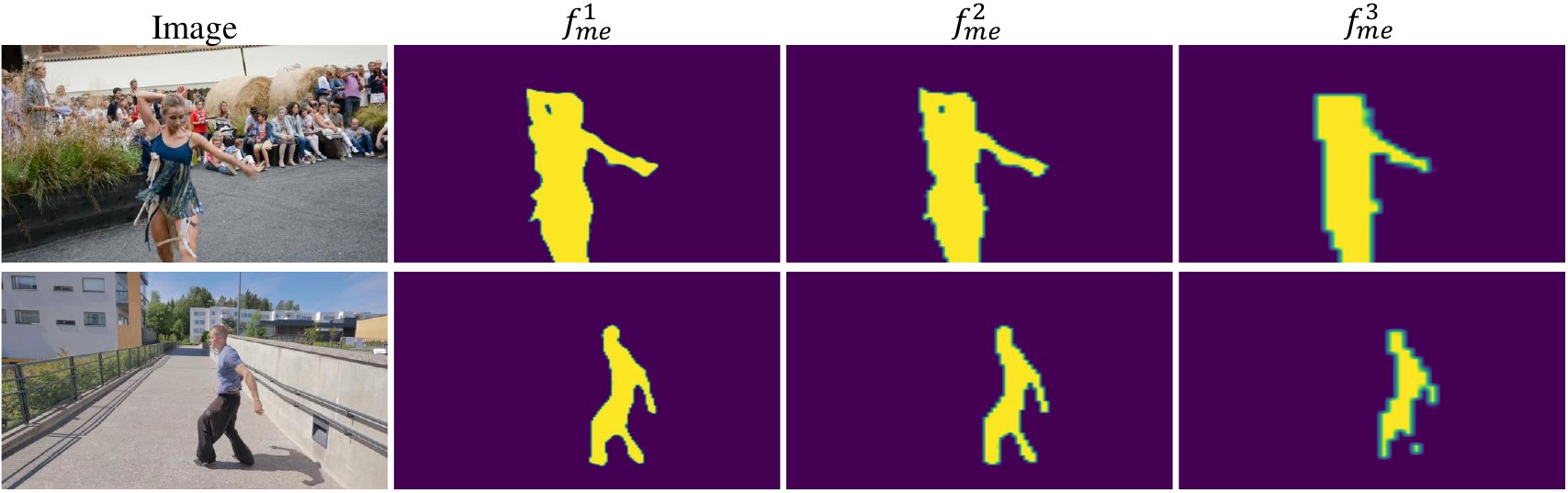}\\
        \centering
        \caption{Visualization of the motion enhance maps ${A}_{me}^{1}$, ${A}_{me}^{2}$ and ${A}_{me}^{3}$. Samples $dance-twirl$ and $parkour$ are randomly selected from the DAVIS$_{16}$~\cite{davis16}.}
\label{fig:Figure10}
% \vspace{-5mm}
\end{figure}

\begin{figure}[!t]
\includegraphics[width=\linewidth]{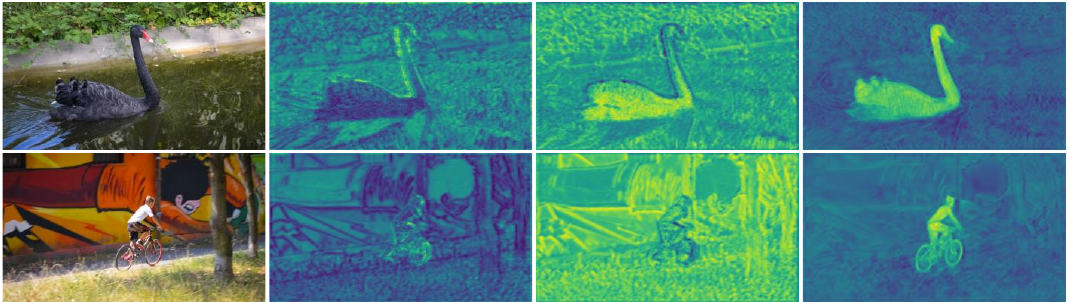}\\
        \centering
        \caption{Visual results of ${P}_{comm}^1$, ${P}_{exclu}^1$ and ${P}^1$ for showing the  effect of the feature purification module. Samples $blackswan$ and $bmx-trees$ are randomly selected from the DAVIS$_{16}$~\cite{davis16}.}
\label{fig:Figure11}
\end{figure}
\begin{figure*}[!t]
    \includegraphics[width=\linewidth]{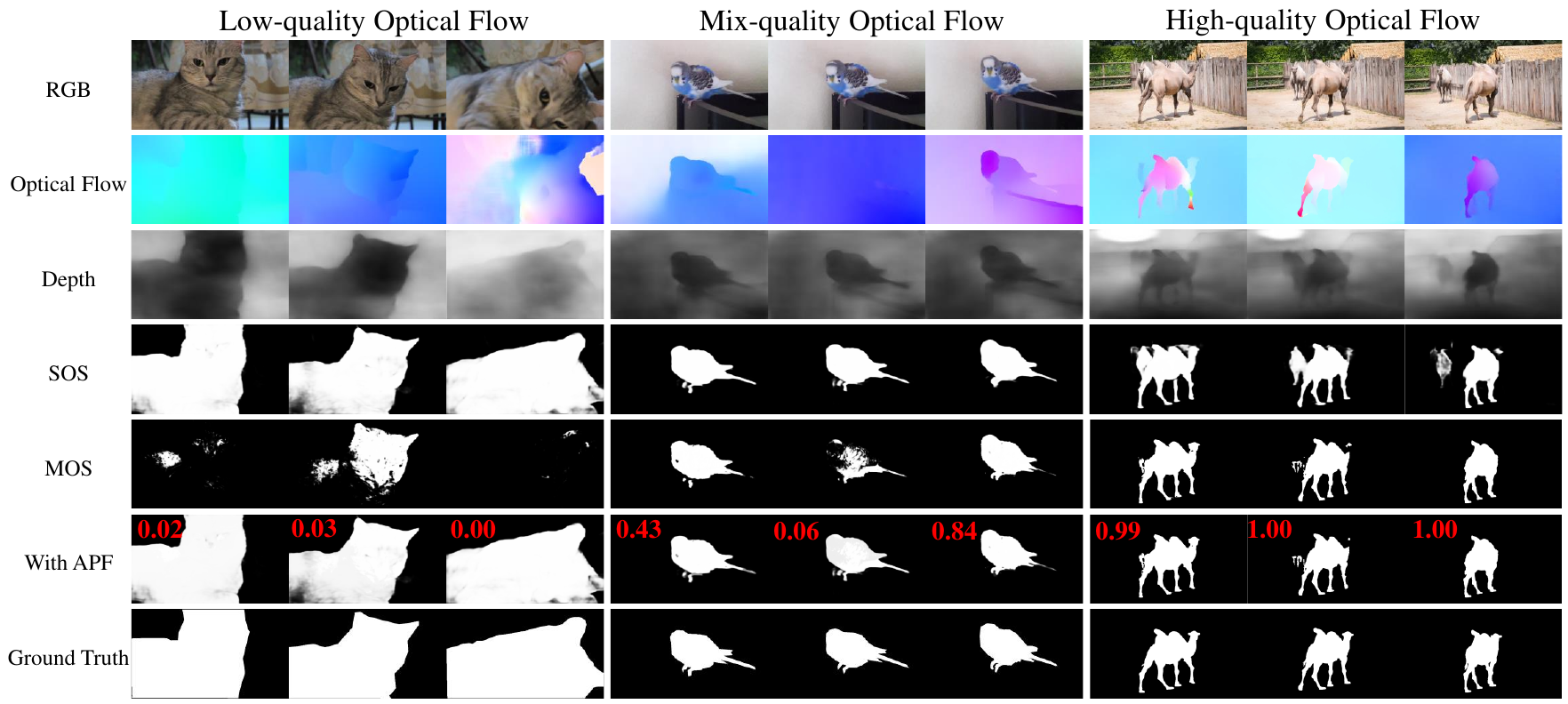}\\ %,height=0.3\linewidth
    \centering
    \caption{Qualitative results on three example videos: $cat$ (low-quality optical flow) and $bird$ (mix-quality optical flow) are from Youtube-Objects, $camel$ (high-quality optical flow) is from DAVIS$_{16}$.} 		
    \label{fig:Figure12}
    % \vspace{-5mm}
\end{figure*} 
\subsubsection{Effectiveness of Multi-source Fusion}\label{subsubsec:multi-source-fusion}
We verify the effectiveness of ISAM, MEM and FPM in multi-source feature fusion, as shown in Tab.~\ref{tab:Table4}.
Compared to directly concatenating multi-source feature maps, the ISAM achieves an improvement of $0.9$\% and $1.5$\% in terms of mean $\mathcal{J}$ and mean $\mathcal{F}$. With the application of MEM, it can obtain the gain of $2.4$\% and $3.1$\% in terms of mean $\mathcal{J}$ and $\mathcal{F}$. By further equipping the FPM, it can yield continuous performance improvement of $1.2$\% and $1.9$\% in terms of mean $\mathcal{J}$ and $\mathcal{F}$, respectively (finally obtaining a total of $86.3$ and $87.3$, respectively). 
In addition, the ISAM-like multi-source interaction module also plays an important role in the static predictor based on the multi-task mechanism, as shown in Tab.~\ref{tab:Table4}. 
The performance gap between the ``+ ISAM'' model and the ``Baseline'' model in the static predictor indicates the necessity of information interaction among different types of features. 
{To further show the advantages of ISAM and MEM in terms of efficiency and accuracy, we replace them with popular transformer~\cite{transformer} modules which utilize multi-head attention with positional encoding to preserve spatial information. As shown in Tab.~\ref{tab:Table4_1}, ISAM and MEM require less than 30\% training memory and 40\% parameters of the transformer components to produce performance improvements of $0.8$\% and $1.8$\% in terms of mean $\mathcal{J}$ and $\mathcal{F}$.}

\begin{table}[t]
  \centering
  \caption{Evaluation of the APF network on both DAVIS$_{16}$ and Youtube-Objects in terms of mean $\mathcal{J}$. {Ideal} denotes the ideal performance of {APF}.}
  \begin{threeparttable}
   \resizebox{\linewidth}{!}{
    \setlength\tabcolsep{7pt}
    \renewcommand\arraystretch{1}
   \input{table/aps_ablation_study_vos}
}
\end{threeparttable}
\label{tab:Table5}
\end{table}

Fig.~\ref{fig:Figure9} $\sim$ ~\ref{fig:Figure11}  shows visual results of the ablation study. 
In Fig.~\ref{fig:Figure9}, we first can see that a large amount of background noise is suppressed in the enhanced RGB and depth feature maps, and the boundaries of static saliency further obtain high-response attention activation values.
Next, we show the visualization of the attention maps $A_{me}^{i}$ from MEM, as shown in Fig.~\ref{fig:Figure10}. It can be seen that high-level attention maps of each layer have clear motion information guidance, and the low-level ones have sharp boundaries of moving objects.
Finally, the FPM mechanism can be intuitively explained by Fig.~\ref{fig:Figure11}. 
The moving objects in ${P}_{comm}^1$ are disturbed by much background information, whereas ${P}_{exclu}^1$ can perspective the background region well. Their subtraction in the FPM actually builds a kind of information constraint by diverse parameters, thereby guaranteeing that ${P}^1$ can focus on moving objects.

\subsubsection{Effectiveness of Adaptive Predictor Fusion}\label{subsubsec:aps}
% \textbf{Effectiveness of Automatic Predictor Selection.} 
In the prediction of Tab.~\ref{tab:Table5}, we detail the performance comparison of using APF network on the Youtube-Objects and DAVIS$_{16}$ datasets, respectively. 
With the help of high-quality optical flow maps, the MOS evidently outperforms the SOS on the DAVIS$_{16}$. On the contrary, the low-quality optical flow on the Youtube-Objects encumbers the prediction of the MOS, that is, they are not as good as those of SOS. The designed APF network comprehensively evaluates the degree of interference of optical flow to the MOS. 
It can be seen that the APF outperforms either the stand-alone SOS or MOS on both the DAVIS$_{16}$ and Youtube-Objects. 
However, the previous APS can not surpass the SOS on Youtube-Objects, this indicates the superiority of adaptive fusion compared to the hard selection strategy.
We further list the ideal performance of APF. The minor gap (1.19\% Youtube-Objects, 0.11\% DAVIS$_{16}$) between the current APF and the ideal one shows our model achieves high-accuracy predictor evaluation and assigns the suitable weight for prediction fusion.
\begin{table*}[!t]
  \centering
  \caption{Quantitative comparison of different RGB-D SOD methods. $\uparrow$ and $ \downarrow$ indicate that the larger scores and the smaller ones  are better, respectively.  The corresponding used backbone network is listed below each model name. R-50, R-101 and R2-50 are the ResNet-50, ResNet-101 and Res2Net-50, respectively.  ``$*$'' denotes the models are trained on the DUT-RGBD + NJUD + NLPR, the others are trained on NJUD + NLPR. The best result for each metric is highlighted in \textbf{bold}.}
  \begin{threeparttable}
   \resizebox{0.9\linewidth}{!}{
    \setlength\tabcolsep{5pt}
    \renewcommand\arraystretch{1}
   \input{table/rgbd_comparison}
}
\end{threeparttable}
\label{tab:Table6}
\end{table*}

\begin{table*}[!t]
  \centering
  \caption{Quantitative evaluation of applying the APF to existing RGB-D SOD methods.}
  \begin{threeparttable}
   \resizebox{0.8\linewidth}{!}{
    \setlength\tabcolsep{5pt}
    \renewcommand\arraystretch{1}
    \input{table/aps_rgbd_ablation_study}
}
\end{threeparttable}
\label{tab:Table7}
\end{table*}

To more intuitively show the effectiveness of the APF, we visualize all sources and the results of each stage under optical flow maps of various qualities in Fig.~\ref{fig:Figure12}. It can be observed that the APF tends to give low weights to the video sequences with low-quality optical flow and high weights to the video sequences with high-quality optical flow, respectively. 
For some frames with mixed-quality optical flow maps, APF will accordingly predict an intermediate weight to comprehensively fuse the results of SOS and MOS.

\subsection{Performance on RGB-D SOD}\label{subsec:performance_rgbd_sod}
According to whether the depth map is required during the testing phase, RGB-D SOD methods can be divided into depth-based and depth-free styles. {Following most methods~\cite{HDFNet,BBSNet,JL-DCF_RGBDSOD,DCF_RGBDSOD,RD3D_RGBDSOD,DASNet_RGBDSOD,CoNet_RGBDSOD} about the choices of backbones and training sets, we train four different versions under the ResNet-50 and ResNet-101 backbones.}  Tab.~\ref{tab:Table6} shows the quantitative comparison with several previous state-of-the-art methods. 
Our multi-task static predictor outperforms all of the competitors and obtains the best performance in terms of all metrics on the NLPR, SIP and DUTLF-D datasets. 
{Among depth-free methods, ``Ours-R-50'' achieves competitive performance with the ``DASNet-R-50''~\cite{DASNet_RGBDSOD} and ``Ours$*$-R-101'' achieves significant improvement against the second best depth-free (``CoNet$*$-R101~\cite{CoNet}) model.}

\begin{figure*}
    \includegraphics[width=\linewidth]{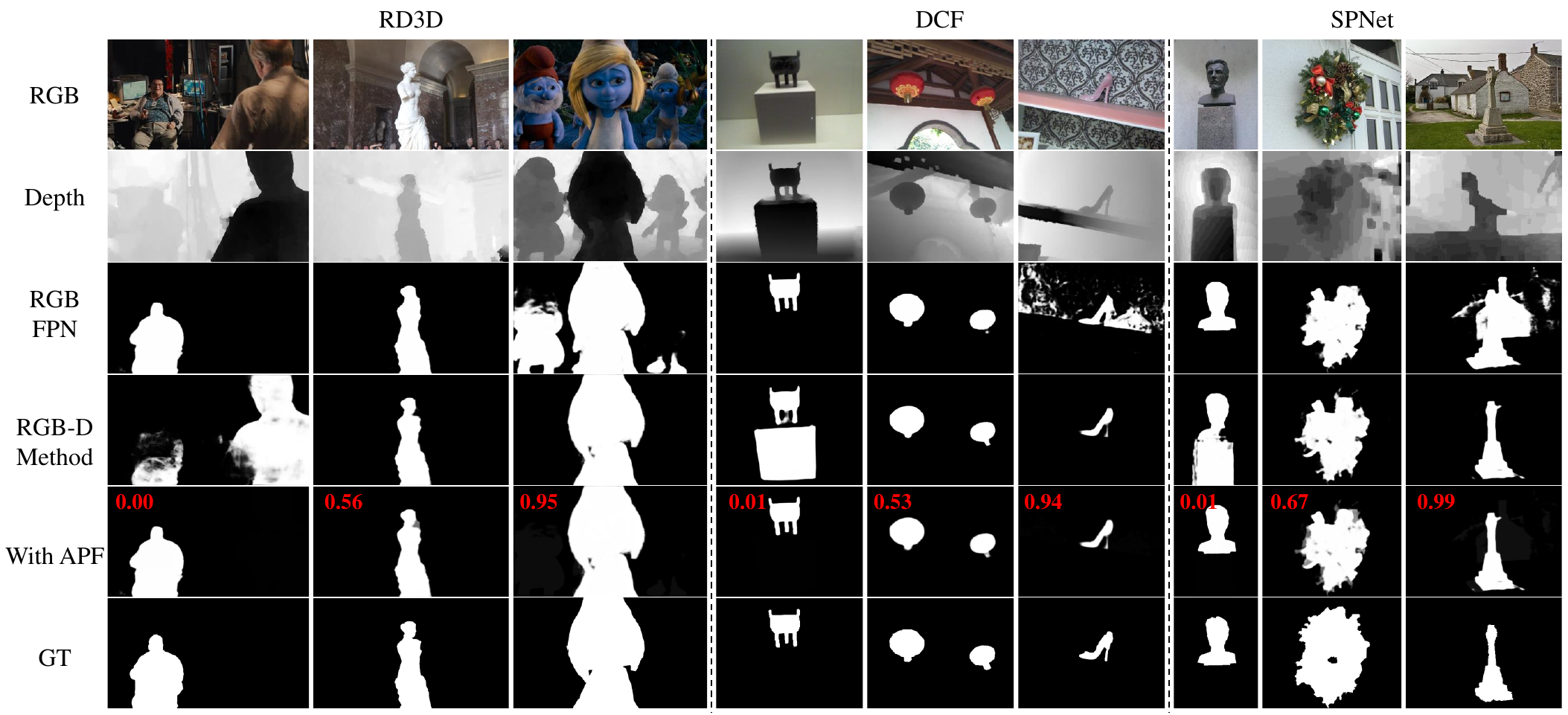}\\ %,height=0.3\linewidth
    \centering
    \caption{Qualitative results of applying APF to previous top-performing RGB-D SOD methods~\cite{RD3D,SPNet,DCF}. The samples for each method are randomly selected from the NJUD~\cite{NJU2K}, NLPR~\cite{NLPR} and STERE~\cite{STERE}, respectively.} 		
    \label{fig:Figure14}
    % \vspace{-5mm}
\end{figure*} 
\begin{table*}[!t]
\large
\centering
	% \scriptsize
	\setlength{\abovecaptionskip}{2pt}
 \caption{Quantitative comparison for depth estimation with two depth-free methods on NLPR, SIP and STERE. The best result for each metric is highlighted in \textbf{bold}.}
 \begin{threeparttable}
   \resizebox{0.7\linewidth}{!}{
    \setlength\tabcolsep{5pt}
    \renewcommand\arraystretch{1}
 \input{table/depth_estimation_comparison}
	}
	\setlength{\abovecaptionskip}{2pt}
	%	\vspace{-5mm}
	\label{tab:Table8}
	\end{threeparttable}
\end{table*}

Besides, we also conduct an extension for evaluating depth quality in RGB-D SOD with the help of our APF network. 
We first construct the FPN-based~\cite{FPN} single stream (RGB-FPN) and the two-stream (RGBD-FPN) network, respectively. 
We choose ResNet-101 as the backbone for both RGB-FPN and RGBD-FPN. Then, we utilize APF to evaluate the quality of the depth map and fuse the predictions from RGB-FPN and RGBD-FPN network. 
Obviously, the baseline RGBD-FPN can be replaced by other existing RGB-D SOD methods. 
Tab.~\ref{tab:Table7} shows the results in terms of three metrics on five RGB-D SOD datasets.
It can be seen that the ``+ APF'' results consistently outperform each pair of RGB and RGB-D predictions under three metrics on the NJUD, STERE, and DUTLF-D datasets. 
Finally, we visualize different RGB-D SOD results shown in Fig.~\ref{fig:Figure14}. It can be seen that the APF models tend to generate low weights for the RGB images with low-adaptation depth maps and high weights for the high-adaptation pairs. 
Although there are some uncomplicated scenes with good depth maps, the APF will predict middle scores because the segmentation results of the RGB-FPN and the RGB-D method are similar (see the $2^{nd}$, $5^{th}$ and $8^{th}$ columns).

\begin{figure*}
    \includegraphics[width=0.8\linewidth]{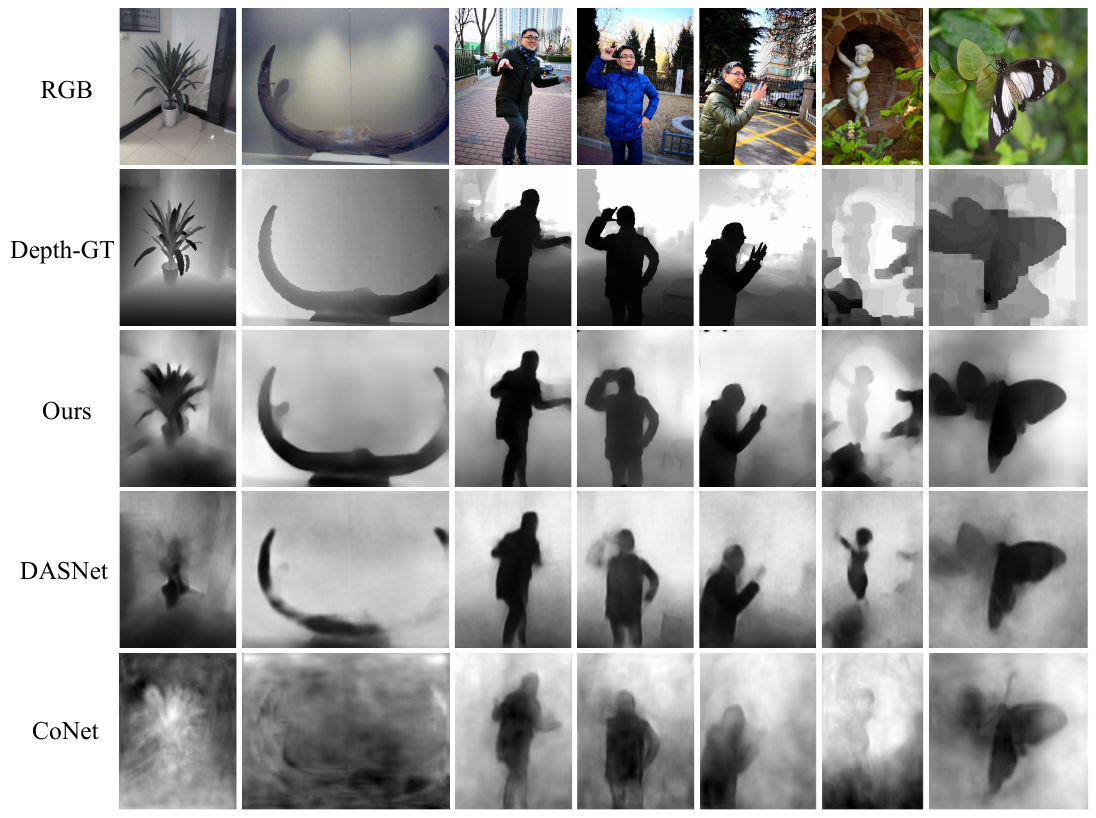}\\ %,height=0.3\linewidth
    \centering
    \caption{Visual comparisons with the DASNet~\cite{DASNet_RGBDSOD} and CoNet~\cite{CoNet_RGBDSOD} in the depth estimation task. Samples are randomly selected from the NLPR~\cite{NJU2K} (1$^{st}$ - 2$^{nd}$ columns ), SIP~\cite{NLPR} (3$^{rd}$ - 5$^{th}$ columns) and STERE~\cite{STERE} (6$^{th}$ - 7$^{th}$ columns).} 	
    \label{fig:Figure15}
    % \vspace{-5mm}
\end{figure*}

\subsection{Performance on Depth Estimation} \label{subsec:performance_depth} 
Tab.~\ref{tab:Table8} shows the results of depth estimation on the NLPR, SIP and STERE datasets in terms of seven common metrics.
It can be seen that the proposed method outperforms other two depth-free RGB-D SOD models across all seven metrics. 
In order to fully demonstrate our superiority, we visualize some results in Fig.~\ref{fig:Figure15}. Benefiting from the proposed ISAM-like multi-task interaction design, we obtain better visual depth predictions with low noise, smooth body and sharp contour.

\section{Conclusion}\label{sec5}
In this paper, we propose novel multi-source static and moving object predictors to effectively utilize the complementary features from the RGB, depth, static saliency and optical flow for zero-shot video object segmentation. 
The proposed interoceptive spatial attention module can leverage complementary information from adjacent sources features to enhance the spatial awareness of each current source features while maintaining source-specific characteristics.
We put forward a motion-enhanced module to comprehensively improve the optical flow features representation by applying the spatial and channel attention technique.
With the help of the feature purification module, it can further filter the incompatible features among sources to refine the multi-source fused features. 
In addition, to get rid of the inevitable interference caused by low-quality optical flow, we design a novel adaptive predictor fusion network, 
which fuses the results from static predictor and moving object predictor according to the generated adaptive weights. The adaptive predictor fusion network is simple yet effective, therefore, it can provide an important reference for other optical flow-based methods and this quality evaluation spirit can be carried towards other auxiliary cues, such as depth maps.
Experiments on three ZVOS datasets indicate that the proposed three-stage method performs favorably against the current state-of-the-art methods. 
Extended experiments on five RGB-D SOD datasets show that the multi-task static predictor performs well in both salient object detection and depth estimation tasks.
\\[3pt] 
\noindent{\textbf{Acknowledgements.}}
Xiaoqi Zhao and Shijie Chang contributed equally to this work. This work was supported by the National Key R\&D Program of China \#2018AAA0102001 and the National Natural Science Foundation of China \#62276046. 
%%===========================================================================================%%
%% If you are submitting to one of the Nature Portfolio journals, using the eJP submission   %%
%% system, please include the references within the manuscript file itself. You may do this  %%
%% by copying the reference list from your .bbl file, paste it into the main manuscript .tex %%
%% file, and delete the associated \verb+\bibliography+ commands.                            %%
%%===========================================================================================%%
{
\small
\bibliographystyle{plain}
\bibliography{sn-bibliography}% common bib file
}
% \bibliography{sn-bibliography}% common bib file
%% if required, the content of .bbl file can be included here once bbl is generated
%%\input sn-article.bbl

%% Default %%
%%\input sn-sample-bib.tex%
\end{document}

%% file: table/scales_training.tex
\begin{tabular}{r||c||c}
			\hline
			{Methods} &	{Video Data} &{Image Data}\\
			\hline
			\hline %\hline \hline
			{MATNet~\cite{MATNet}} &$\sim$ 14,000&-\\
			{AGS~\cite{AGS}}  &$\sim$ 8,500&$\sim$ 6,000 \\
              {COSNet~\cite{COSNet}}  &$\sim$ 2,000&$\sim$ 15,000\\
	      {AGNN~\cite{AGNN}}  &$\sim$ 2,000 &$\sim$ 15,000\\
	 {GateNet~\cite{GateNet}}  &$\sim$ 2,000 &$\sim$ 15,000\\
   {FSNet~\cite{FSNet}}  &$\sim$ 2,000 &$\sim$ 10,000\\
    {AMCNet~\cite{AMCNet}}  &$\sim$ 2,000 &$\sim$ 10,000\\
     {RTNet~\cite{RTNet}}  &$\sim$ 2,000 &$\sim$ 10,000\\
      {Ours}  &$\sim$ 2,000 &$\sim$ 6,300\\
			\hline
	\end{tabular}

%% file: table/davis16_comparison.tex
\begin{tabular}{cr||cccccccc||ccccccccc}
	   \toprule%[1pt]
	    & & &\multicolumn{5}{c}{Interframe-based methods}&&&\multicolumn{5}{c}{Optical flow-based methods} \\ \hline
		& Methods  &PDB  &MotAdapt  & EPO &AGS & COSNet  & AGNN & DFNet & WCS & SFL & MP & GateNet& MATNet  & FSNet& AMCNet&RTNet &Ours \\ 
		& &~\cite{PDB}&~\cite{MotAdapt}&~\cite{EPO}&~\cite{AGS}&\cite{COSNet}&~\cite{AGNN}&~\cite{DFNet}&~\cite{WCS}&~\cite{SFL}&~\cite{MP}&~\cite{GateNet}&~\cite{MATNet}&~\cite{FSNet}&~\cite{AMCNet}&~\cite{RTNet}&ours\\
		\hline \hline
		& mean $\mathcal{J}$ $\uparrow$   &77.2  &77.2  &80.6  &79.7  &80.5  & 80.7& 80.4 &82.2 & 67.4 & 70.0 & 80.9& 82.4  & 83.4 & 84.5 &85.6 & \textbf{87.1}   \\
		\hline
		& mean $\mathcal{F}$ $\uparrow$   &74.5  &77.4  &75.5  &77.4  &79.5  & 79.1 & -- & 80.7 & 80.7 &66.7&65.9&79.4 & 83.1  & 84.6  &84.7 & \textbf{87.5} \\
 	 \bottomrule[1pt]
	\end{tabular}

%% file: table/youtube_comparison.tex
	\begin{tabular}{r||cccccccccc|c}
		 \toprule%[1pt]
			& Airplane (6) & Bird (6) & Boat (15)  & Car (7) & Cat (16)& Cow (20) & Dog (27) &Horse (14)& Motorbike (10)& Train (5) & Avg. \\
			%\rowcolor{mygray}
			\hline
			\hline
			FST~\cite{FST}& 70.9 &70.6& 42.5&65.2  &52.1&44.5&65.3 &53.5&44.2 &29.6 &53.8  \\		
			COSEG~\cite{COSEG}&69.3   &76.0 &53.5  &70.4 & 66.8& 49.0 & 47.5& 55.7& 39.5& 53.4& 58.1\\
			%\rowcolor{mygray}
			ARP~\cite{ARP} &73.6 & 56.1&57.8 &33.9& 30.5& 41.8& 36.8 &44.3& 48.9& 39.2&46.2\\
			LVO~\cite{LVO} &86.2 &81.0&68.5 &69.3& 58.8&68.5&61.7 &53.9& 60.8& \textbf{66.3}&67.5\\
			%\rowcolor{mygray}
			PDB~\cite{PDB} &78.0  &80.0 &58.9 &76.5 &63.0&64.1&70.1 &67.6&58.3&35.2&65.4\\
			FSEG~\cite{FSEG}  &{81.7}  &63.8 &{72.3}&74.9&68.4&68.0 &69.4 & 60.4&62.7&62.2&68.4\\
			%\rowcolor{mygray}
			AGS~\cite{AGS} &\textbf{87.7}&76.7&72.2&78.6&69.2&64.6&73.3&64.4&62.1&48.2&69.7\\
			COSNet~\cite{COSNet} &81.1&75.7&71.3&77.6&66.5&{69.8}&76.8&67.4&67.7&46.8&70.5\\
			AGNN~\cite{AGNN} &81.1  &75.9 &70.7 &78.1 &67.9&69.7&{77.4} &67.3&68.3&47.8&70.8\\
			WCS~\cite{WCS} &81.8  &81.2 &67.6 &79.5 &65.8&66.2&73.4 &\textbf{69.5}&\textbf{69.3}&49.7&70.9\\
			\hline\hline
			SFL~\cite{SFL}&65.6  &65.4 &59.9&64.0 &58.9&51.1& 54.1 &64.8& 52.6& 34.0&57.0\\
			MATNet~\cite{MATNet} &72.9  &77.5 &66.9 &79.0 &73.7&67.4&75.9 &63.2&62.6&51.0&69.0\\
			AMCNet~\cite{AMCNet} &78.9  &80.9 &67.4 &82.0 &69.0&69.6&75.8 &63.0&63.4&57.8&71.1\\
			RTNet~\cite{RTNet} &84.1  &80.2 &70.1 &79.5 &71.8&70.1&71.3 &65.1&64.6&53.3&71.0\\
			Ours &86.1  &\textbf{82.4} &\textbf{73.0} &\textbf{79.9} &\textbf{80.0} &\textbf{72.6} &\textbf{80.3}&66.5&66.4&62.6&\textbf{75.0}\\
		\bottomrule[1pt]
    \end{tabular}

%% file: table/fbms_comparison.tex
\begin{tabular}{c||cccccccc}
	\hline
	Methods	& ARP& SAGE & LVO& FSEG & PDB & MATNet&AMCNet&Ours\\ 
	&~\cite{ARP}&~\cite{SAGE}&~\cite{LVO}&~\cite{FSEG}&~\cite{PDB}&~\cite{MATNet}&~\cite{AMCNet}&ours\\\hline
	mean $\mathcal{J}\uparrow$ & 59.8 & 61.2& 65.1& 68.4& 74.0& 76.1&76.5&\textbf{81.3}\\
	\hline
	\end{tabular}

%% file: table/ablation_study_davis16.tex
\begin{tabular}{c|c||cc}
			\hline
			Components &Module &mean $\mathcal{J}$ $\uparrow$&mean $\mathcal{F}$ $\uparrow$\\
			\hline
			\hline %\hline \hline
			\multirow{2}{*}{{Static predictor}}&{Baseline} &76.4 &75.8\\
			&{+ISAM} &76.7 &76.8\\
			\hline
            \hline
			\multirow{5}{*}{{Multi-source}}&{RGB} &77.7 &77.6\\
			&{RGB+D} &79.4 &78.5\\
			&{RGB+SOS} &79.4 &79.0\\
	       {feature inputs} &{RGB+OF} &80.2 &79.2\\
	        &{RGB+D+SOS+OF} &82.3 &81.7\\
			\hline
            \hline
			\multirow{3}{*}{{Multi-source  fusion}} & +ISAM & 83.1 & 82.9\\
			& +MEM & 85.1 & 85.5\\
			& +FPM & 86.3 & 87.3\\
			%&Prediction segmentation fusion  & 79.4 & -1.2 & 74.2 & -2.4 \\
            \hline
		    \hline
			Prediction& +APF  & 87.1 & 87.5\\
			\hline
	\end{tabular}

%% file: table/ablation_study_davis16_replace_transformer.tex
\begin{tabular}{cc|cc||cccc}
			\hline
			ISAM&Transf. &MEM &Transf.&Memory $\downarrow$&Params $\downarrow$&mean $\mathcal{J}$ $\uparrow$ &mean $\mathcal{F}$ $\uparrow$\\
			\hline %\hline \hline
			&\checkmark&&\checkmark &31GB &31.7MB&85.6 &85.8\\
   	      &\checkmark&\checkmark& &19GB &23.3MB&85.6 &86.6\\
		    \checkmark&&&\checkmark &22GB &20.8MB&86.3 &86.7\\
                \checkmark&&\checkmark &&\textbf{10GB} &\textbf{12.3MB}&\textbf{86.3} &\textbf{87.3}\\
			\hline
	\end{tabular}

%% file: table/aps_ablation_study_vos.tex
\begin{tabular}{c||c||c}
			\hline
			{Prediction} &	{Youtube-Objects} &{DAVIS$_{16}$}\\
			\hline
			\hline %\hline \hline
			{SOS} &74.0 ($\downarrow$2.50\%)&76.7 ($\downarrow$12.04\%)\\
			{MOS}  &65.0 ($\downarrow$14.36\%)&86.3 ($\downarrow$1.03\%)\\
              {APS}  &70.0 ($\downarrow$7.77\%)&86.8 ($\downarrow$0.46\%)\\
	        {APF}  &75.0 ($\downarrow$1.19\%)&87.1 ($\downarrow$0.11\%)\\
	        {Ideal}  &75.9  &87.2\\
			\hline
	\end{tabular}

%% file: table/rgbd_comparison.tex
 
 \begin{tabular}{c|c||ccccccc||cccccc}
	\toprule
	& & \multicolumn{7}{c||}{\textbf{Depth-based methods}} & \multicolumn{6}{c}{\textbf{Depth-free methods}} \\
	\hline
	\multirow{3}{*}{} & \multirow{3}{*}{Metric} & HDFNet & BBSNet & JL-DCF &  DSA2F$*$ & DCF$*$ & RD3D & SPNet & DASNet & CoNet$*$ & Ours$*$ & Ours & Ours$*$ & Ours  \\
				& & \cite{HDFNet} & \cite{BBSNet} & \cite{JLDCF} & \cite{D2F}  & \cite{DCF}  & \cite{RD3D} & \cite{SPNet}& \cite{DASNet} & \cite{CoNet} & &  &  &  \\ 
    	& & R-50 & R-50 & R-101 & VGG-19 & R-50 & R-50 & R2-50 &  R-50 &  R-101 & R-50 & R-50 & R-101 & R-101 \\ \hline
				\multirow{4}{*}{\rotatebox{90}{NJUD}~\rotatebox{90}{~\cite{NJU2K}}}    
				& $E_{\xi}\uparrow$ 
				& .932 & \textbf{.942} & .935 & .937 & .941 & \textbf{.942} & .936 &  .936 & .924 & .934 & .931 & .938 & .937 \\
				& $S_{\alpha}\uparrow$ 
				& .908 & .921 & .902 & .904 & .903 & .916 & .925 & .902 & .894 & .919 & .915 & \textbf{.927} & .926 \\
				& $F_{\beta}\uparrow$ 
				& .922 & .931 & .912 & .917 & .917 & .923 & .934 & .911 & .902 & .933 & .929 & \textbf{.941} & .940 \\
				& $M\,\downarrow$ 
				& .038 & .035 & .041 & .039 & .038 & .037 & .030 & .042 & .047 & .034 & .035 & \textbf{.029} & .030 \\ \hline
				
				\multirow{4}{*}{\rotatebox{90}{NLPR}~\rotatebox{90}{~\cite{NLPR}}}    
				& $E_{\xi}\uparrow$ 
				& .957 & .954 & .955 & .952 & .956 & .959 & .959 & .961 & .934 & .953 & .957 & .959 & \textbf{.964} \\
				& $S_{\alpha}\uparrow$ 
				& .923 & .930 & .925 & .918 & .921 & .929 & .928 & .929 & .907 & .925 & .922 & .931 & \textbf{.932} \\
				& $F_{\beta}\uparrow$ 
				& .927 & .927 & .925 & .916 & .917 & .927 & .925 & .929 & .898 & .923 & .924 & .931 & \textbf{.936} \\
				& $M\,\downarrow$ 
				& .023 & .023 & .022 & .024 & .023 & .022 & .022 & .021 & .031 & .024 & .023 & .021 & \textbf{.020} \\ \hline
				
				\multirow{4}{*}{\rotatebox{90}{STERE}~\rotatebox{90}{~~\cite{STERE}}}    
				& $E_{\xi}\uparrow$ 
				& .931 & .941 & .937 & .942 & .943 & \textbf{.944} & .940 & .941 & .923 & .936 & .937 & .937 & .938 \\
				& $S_{\alpha}\uparrow$ 
				& .900 & .908 & .903 & .897 & .905 & .911 & .907 & .910 & .908 & .917 & .914 & \textbf{.921} & \textbf{.921} \\
				& $F_{\beta}\uparrow$ 
				& .910 & .919 & .913 & .910 & .915 & .917 & .913 & .915 & .909 & .920 & .920 & .923 & \textbf{.925} \\
				& $M\,\downarrow$ 
				& .041 & .041 & .040 & .039 & .037 & .037 & .038 & .037 & .041 & .033 & .034 & \textbf{.031} & \textbf{.031} \\ \hline
				
%				\multirow{4}{*}{\rotatebox{90}{DES}~\rotatebox{90}{~\cite{cheng2014depth}}}    
%				& $E_{\xi}\uparrow$ 
%				& .971 & .966 & .969 & .962 & - & .931 & \textbf{.983} & .942 & .945 & .953 & .949 & .969 & .954 \\
%				& $S_{\alpha}\uparrow$ 
%				& .926 & .933 & .931 & .920 & - & .873 & \textbf{.946} & .905 & .910 & .909 & .908 & .926 & .916 \\
%				& $F_{\beta}\uparrow$ 
%				& .932 & .927 & .934 & .896 & - & .894 & \textbf{.952} & .927 & .861 & .920 & .922 & .937 & .932 \\
%				& $M\,\downarrow$ 
%				& .021 & .021 & .021 & .021 & - & .037 & \textbf{.014} & .025 & .027 & .025 & .024 & .019 & .022 \\ \hline
				
				\multirow{4}{*}{\rotatebox{90}{SIP}~\rotatebox{90}{~\cite{SIP}}}    
				& $E_{\xi}\uparrow$ 
				& .925 & .917 & .923 & .911 & .921 & .924 & .931 & .923 & .909 & .932 & .922 & \textbf{.936} & .932 \\
				& $S_{\alpha}\uparrow$ 
				& .886 & .879 & .880 & .862 & .873 & .885 & .896 & .877 & .858 & .896 & .883 & \textbf{.899} & .896 \\
				& $F_{\beta}\uparrow$ 
				& .910 & .902 & .904 & .891 & .900 & .906 & .916 & .900 & .883 & .918 & .907 & \textbf{.922} & .920 \\
				& $M\,\downarrow$ 
				& .047 & .055 & .049 & .057 & .052 & .048 & .043 & .051 & .063 & .041 & .047 & \textbf{.039} & .041 \\ 
				\hline
				
				\multirow{4}{*}{\rotatebox{90}{DUTLF}~\rotatebox{90}{-D \cite{DUTLF-D}}}
				& $E_{\xi}\uparrow$ 
				& .938 & - & .938 & .956 & .957 & .957 & - & - & .953 & .955 & - & \textbf{.963} & - \\
				& $S_{\alpha}\uparrow$ 
				& .907 & - & .905 & .921 & .924 & .931 & - & - & .919 & .930 & - & \textbf{.937} & - \\
				& $F_{\beta}\uparrow$ 
				& .930 & - & .924 & .938 & .940 & .947 & - & - & .935 & .944 & - & \textbf{.951} & - \\
				& $M\,\downarrow$ 
				& .941 & - & .043 & .031 & .030 & .031 & - & - & .033 & .028 & - & \textbf{.026} & - \\ 
				\hline
	\end{tabular}

%% file: table/aps_rgbd_ablation_study.tex
\begin{tabular}{c||ccc|ccc|ccc|ccc|ccc}
\toprule
 Dataset & \multicolumn{3}{c|}{NJUD} & \multicolumn{3}{c|}{NLPR} & \multicolumn{3}{c|}{STERE} & \multicolumn{3}{c|}{SIP} & \multicolumn{3}{c}{DUTLF-D}\\
 \hline
 Metrics & $E_{\xi}\uparrow$ & $S_{\alpha}\uparrow$ & $F_{\beta}\uparrow$ & $E_{\xi}\uparrow$ & $S_{\alpha}\uparrow$ & $F_{\beta}\uparrow$ & $E_{\xi}\uparrow$ & $S_{\alpha}\uparrow$ & $F_{\beta}\uparrow$ & $E_{\xi}\uparrow$ & $S_{\alpha}\uparrow$ & $F_{\beta}\uparrow$ &
 $E_{\xi}\uparrow$ & $S_{\alpha}\uparrow$ & $F_{\beta}\uparrow$ \\
 \hline
 \hline
 RGB-FPN & .949 & .912 & .924 & .954 & .921 & .918 & .944 & .910 & .914 & .924 & .881 & .906 & \textbf{.951} & \textbf{.918} & \textbf{.936} \\
 RGBD-FPN& .951 & .919 & .929 & \textbf{.963} & .925 & .923 & .946 & .908 & .915 & .928 & .890 & .906 & .942 & .902 & .925 \\
 + APF & \textbf{.953} & \textbf{.922} & \textbf{.934} & .961 & \textbf{.927} & \textbf{.924} & \textbf{.948} & \textbf{.915} & \textbf{.922} & \textbf{.928} & \textbf{.891} & \textbf{.912} & \textbf{.951} & \textbf{.918} & \textbf{.936} \\
\hline \hline
 RGB-FPN   & .949 & .912 & .924 & .954 & .921 & .918 & .944 & .910 & .914 & .924 & .881 & .906 & - & - & - \\
 SPNet \cite{SPNet} & \textbf{.956} & .925 & .934 & \textbf{.961} & .928 & .925 & .948 & .907 & .913 & \textbf{.934} & \textbf{.895} & .916 & - & - & - \\
 + APF   & \textbf{.956} & \textbf{.926} & \textbf{.939} & \textbf{.961} & \textbf{.931} & \textbf{.930} & \textbf{.950} & \textbf{.918} & \textbf{.925} & .932 & \textbf{.895} & \textbf{.918} & - & - & - \\
\hline \hline
 RGB-FPN   & .949 & .912 & .924 & .954 & .921 & .918 & .944 & .910 & .914 & .924 & .881 & .906 & .951 & .918 & .936 \\
 RD3D \cite{RD3D}  & .947 & .916 & .923 & \textbf{.965} & .929 & .927 & .947 & .911 & .917 & .924 & .885 & .906 & .960 & \textbf{.931} & .947 \\
 + APF   & \textbf{.954} & \textbf{.923} & \textbf{.935} & \textbf{.965} & \textbf{.933} & \textbf{.933} & \textbf{.948} & \textbf{.918} & \textbf{.927} & \textbf{.929} & \textbf{.891} & \textbf{.916} & \textbf{.961} & \textbf{.931} & \textbf{.951} \\

\hline \hline
 RGB-FPN   & \textbf{.949} & .912 & .924 & .954 & .921 & .918 & .944 & .910 & .914 & \textbf{.924} & \textbf{.881} & .906 & .951 & .918 & .936 \\
 DCF \cite{DCF}  & .943 & .903 & .917 & .957 & .921 & .917 & .948 & .905 & .915 & .922 & .873 & .900 & .957 & .924 & .940 \\
 + APF   & \textbf{.949} & \textbf{.916} & \textbf{.929} & \textbf{.959} & \textbf{.928} & \textbf{.927} & \textbf{.950} & \textbf{.917} & \textbf{.926} & \textbf{.924} & \textbf{.881} & \textbf{.907} & \textbf{.958} & \textbf{.928} & \textbf{.947} \\
\hline
\end{tabular}

%% file: table/depth_estimation_comparison.tex
\begin{tabular}{c|c||cccc|ccc}
\toprule
Dataset                & Methods & Abs Rel $\downarrow$ & Sq Rel $\downarrow$ & RMSE $\downarrow$ & RMSE log $\downarrow$ & P1 $\uparrow$ & P2 $\uparrow$ & P3 $\uparrow$ \\
\hline   
\hline
\multirow{3}{*}{NLPR}  & DASNet \cite{DASNet}   & 0.177 & 0.101 & 0.398 & 0.063 & 0.681 & 0.907  & 0.982 \\  & CoNet \cite{CoNet}   & 0.407 & 0.381 & 0.771 & 0.213 & 0.302 & 0.586  & 0.837 \\
                       & Ours    & \textbf{0.169} & \textbf{0.092} & \textbf{0.377} & \textbf{0.060} & \textbf{0.693} & \textbf{0.911} & \textbf{0.986} \\
\hline  
\hline
\multirow{3}{*}{SIP}  & DASNet \cite{DASNet}   & 0.177 & 0.101 & 0.398 & 0.063 & 0.739 & 0.920  & 0.984 \\  & CoNet \cite{CoNet}   & 0.149 & 0.094 & 0.433 & 0.062 & 0.677 & 0.907  & 0.984 \\
                       & Ours    & \textbf{0.126} & \textbf{0.072} & \textbf{0.362} & \textbf{0.048} & \textbf{0.746} & \textbf{0.933} & \textbf{0.987} \\
\hline
\hline
\multirow{3}{*}{STERE} & DASNet \cite{DASNet}   & 0.192 & 0.107 & 0.413 & 0.064 & 0.660 & 0.911 & 0.987 \\
                    & CoNet \cite{CoNet} & 0.190 & 0.100 & 0.394 & 0.061 & 0.659 & 0.919 & 0.989 \\
                       & Ours    & \textbf{0.180} & \textbf{0.095} & \textbf{0.381} & \textbf{0.057} & \textbf{0.692} & \textbf{0.926} & \textbf{0.988} \\
\hline
\end{tabular}